\documentclass[letterpaper]{article} 
\usepackage{aaai2027}  
\nocopyright
\usepackage[hyphens]{url}  
\usepackage{graphicx} 
\usepackage{natbib}  
\usepackage{caption} 
\usepackage{algorithm}
\usepackage{algpseudocode}
\usepackage{amssymb,amsmath} 

\usepackage{xspace}
\newcommand{\method}{RoomWright\xspace}   

\usepackage{newfloat}
\usepackage{listings}
\DeclareCaptionStyle{ruled}{labelfont=normalfont,labelsep=colon,strut=off} 
\floatstyle{ruled}
\newfloat{listing}{tb}{lst}{}
\floatname{listing}{Listing}

\usepackage{booktabs}

\title{Beyond Placement and Articulation: Usage-Driven Code Scenes for Embodied Interaction}

\author{
    Zijian Xiao\textsuperscript{\rm 1\rm 2}\thanks{Work done during an internship at Meituan.},
    Zipeng Ye\textsuperscript{\rm 2},
    Jinkun Hao\textsuperscript{\rm 1},
    Xiong Yang\textsuperscript{\rm 1\rm 2}\footnotemark[1],
    Yuchen Xie\textsuperscript{\rm 2},
    Ran Yi\textsuperscript{\rm 1}\thanks{Corresponding author}\\
}
\affiliations{
    \textsuperscript{\rm 1}Shanghai Jiao Tong University\\
    \textsuperscript{\rm 2}Meituan
}

\begin{document}

\maketitle

\begin{abstract}

Indoor scene synthesis provides essential environments for embodied AI, robotic manipulation, and simulation-based policy learning. Recent code-based scene generation methods produce editable and extensible environments, yet they remain focused on visual construction and object-level articulation, leaving the functional usage of scenes largely unmodeled. To address this problem, we present \method, an agentic usage-driven framework for generating 3D scenes represented entirely as code for embodied interaction. \method performs usage-driven object reasoning, which treats each anchor as a task centre and admits task-required objects and their affordances. A code agent further enables multi-part interaction by compiling each interaction into a trigger–condition–effect rule that updates structured object states, capturing causal dependencies across objects. Moreover, since manipuland orientation is ambiguous and hard to recover from pixels, \method alleviates this via annotation-informed usage-guided orientation. Extensive experiments demonstrate the effectiveness of our method. The resulting scenes are executable, editable, and simulation-ready, providing interactive environments for embodied AI and policy learning. 

\end{abstract}

\section{Introduction}

Indoor scene synthesis provides essential environments for embodied AI, robotic manipulation, and simulation-based policy learning. Conventional methods typically generate or retrieve individual 3D assets and then arrange them into a scene. Although these approaches can produce visually realistic environments, the resulting scenes are usually represented as static geometry, making their structure, semantics, and behaviors difficult to inspect, modify, and extend.

Recent advances in the coding capabilities of large language models and VLM (Vision Language Models) suggest a new paradigm in which an entire 3D scene is represented as executable code. Code-based representations are highly flexible and preserve explicit modeling procedures, semantic structures, and construction histories. They can be executed for verification, edited at test time, and iteratively extended without rebuilding the scene from scratch. These properties make code a promising representation not only for procedural asset generation, but also for complete and interactive 3D environments.

\begin{figure}[!t]
\centering
\includegraphics[width=1\linewidth]{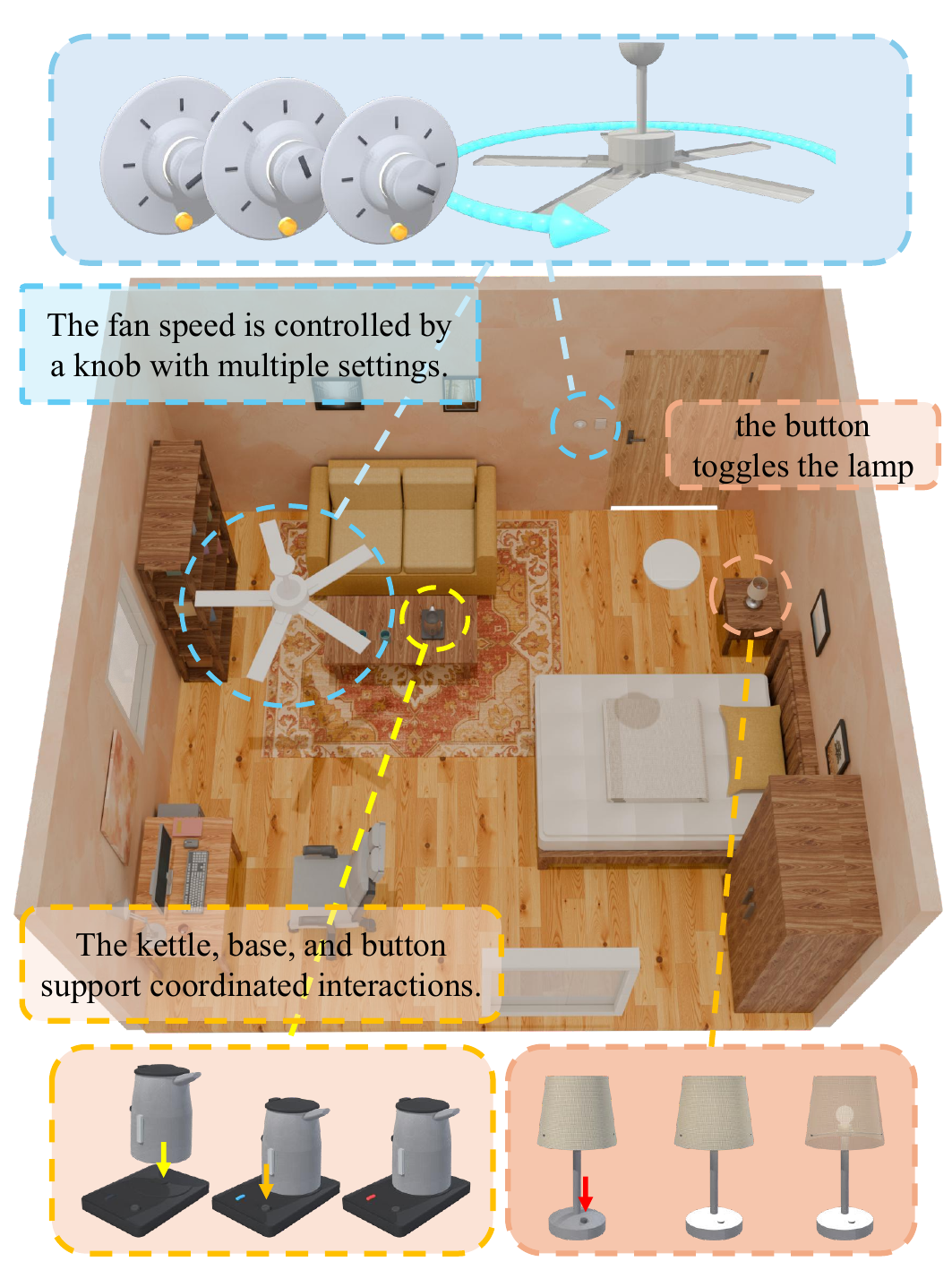}
\caption{
The scenes automatically generated by \method. Each object in the scene is independently separable and possesses its own articulation structure, while different parts of an object, and even multiple objects, exhibit rich coordinated interactions. The objects, the scene layout, and all interaction behaviors are represented entirely as code.
}
\label{fig_teaser}
\end{figure}

Recent works SceneCode~\cite{wang2026scenecode} and Code-as-Room~\cite{yang2026code} have begun to explore the generation of scenes represented entirely as code. They use VLM to evaluate rendered images, generate procedural assets, and refine object placement. SceneCode also exploits code to assign articulation attributes, allowing objects such as doors, windows, and cabinets to open, close, or slide. However, these studies remain primarily focused on visual scene construction and object-level articulation. We argue that code-represented scenes should go beyond placement and articulation, and should instead model how a scene is actually used. 

We present \method, a usage-driven framework for generating code-represented 3D scenes for embodied interaction. Our key insight is to preserve semantic annotations during asset generation and use them to support scene-level reasoning: the language model performs semantic and task reasoning, while deterministic geometry and physics modules compute coordinates, orientations, and executable behaviors. \textbf{(i) Usage-driven object reasoning and user-centered composition.} Existing methods do not explicitly capture this usage structure. They largely rely on the visual judgment of a vision language model. However, a functional scene is not simply a collection of objects that appears visually plausible. Object selection, placement, orientation, and interaction are closely determined by intended usage. When people search for an object, they often recall the activity in which it was last used. \method performs \emph{usage-driven object reasoning and user-centered composition}: given an intended activity, the system infers the required objects and their functional relationships, and represents scene organization using user-relative relations, functional zones, and pairwise spatial constraints, which a geometric solver then converts into collision-free metric placements, allowing the language model to reason about usage while explicit geometry handles numerical coordinates. 

\textbf{(ii) Code-based multi-part interaction.} Current interactive scene generation stops at articulation, which specifies how an individual object part can move, such as a door rotating around a hinge; embodied environments, however, often require causal interactions involving multiple parts or multiple objects, such as pressing a switch to turn on a lamp, turning a knob to different positions to control the speed of a fan, or placing a kettle on a charging base to activate an indicator (Figure \ref{fig_teaser}), which cannot be represented by isolated articulation parameters alone. In \method, a code agent enables \emph{multi-part interaction} by compiling each interaction into a  trigger–condition–effect rule over structured object states, supporting coordinated motion among multiple parts, causal dependencies across different objects, and changes in physical or visual states.

\textbf{(iii) Annotation-informed usage-guided orientation.} The orientation of manipulable objects is difficult to determine from rendered appearance, since important interaction features may be small, occluded, or visually similar to other local structures. Moreover, the meaning of an object's front direction depends on its usage. \method adopts \emph{annotation-informed usage-guided orientation}. We leverage the code representation to record semantic annotations during the modeling stage, providing the VLM with information beyond raw pixels so that it can infer the intrinsic properties of objects without relying on visual appearance alone. Based on these annotations, we assign each manipulable object a use-facing direction according to how it is intended to be used, thereby alleviating the orientation problem of manipulands.

In summary, our main contributions are:

\begin{itemize}
\item We propose \method, a code scene generation agentic framework that produces editable and simulation-ready environments for embodied interaction. Extensive experiments demonstrate the effectiveness of our method.

\item We introduce usage-driven object reasoning and user-centered composition, deriving scene content and spatial organization from intended activities and functional relationships.

\item We enable multi-part interaction via a code agent that represents each interaction as a trigger–condition–effect rule over declared object states, capturing causal dependencies beyond single-object articulation.

\item We enable annotation-informed usage-guided orientation, where semantic annotations recorded during modeling guide the assignment of use-facing directions to manipulable objects, alleviating orientation ambiguity.
\end{itemize}

\begin{figure*}[t]
  \centering
   \includegraphics[width=1\linewidth]{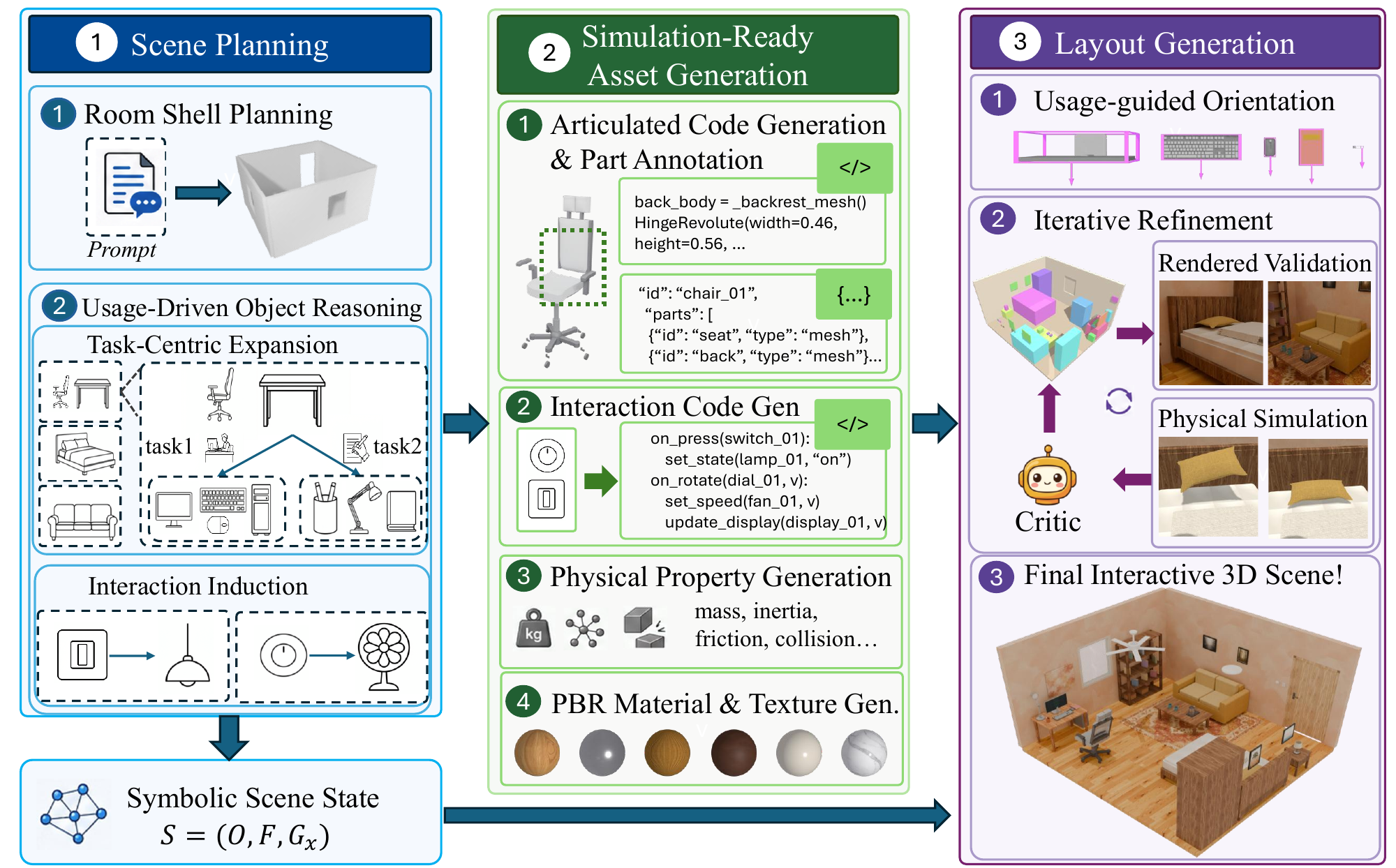}

   \caption{
Pipeline overview of \method.
Starting from a natural-language room description, \method build an interactive 3D scene through three stages:
(1) \textbf{scene planning} infers a usage-driven symbolic scene state with object, support, and interaction relations;
(2) \textbf{simulation-ready asset generation} produces simulation-ready articulated assets with semantic parts, physical properties, and executable interactions;
(3) \textbf{layout generation} optimizes object placement through geometric reasoning, visual critique, and physical simulation.
The generated scene is directly executable in embodied environments.
}
   \label{fig_framework}
\end{figure*}

\section{Related Work}
\subsection{Agentic Scene Synthesis}
Indoor scene synthesis has progressed from procedural and data-driven methods ~\cite{raistrick2024infinigen,paschalidou2021atiss,tang2024diffuscene,zhu2025imaginarium} to open-vocabulary systems that use LLMs or VLMs for layout and scene-graph planning ~\cite{feng2023layoutgpt,yang2024holodeck,ccelen2024design,fu2024anyhome,sun2025layoutvlm,luo2026stable,hao2026mesatask}. Recent agentic methods further introduce iterative planning, tool-using, and feedback. SceneWeaver employs self-reflection, HSM organizes generation hierarchically, and Code-as-Room uses structured execution and persistent memory for image-conditioned room synthesis ~\cite{yang2026sceneweaver,pun2026hsm,yang2026code}. SceneSmith generates simulation-ready environments through designer–critic–orchestrator collaboration, whereas SAGE incorporates visual and physical critics into a task-conditioned generation loop ~\cite{pfaff2026scenesmith,xia2026sage}. These methods improve semantic fidelity and physical validity, but largely focus on scene composition rather than generating functional dependencies among scene entities.
\subsection{Code-Based 3D Generation}
Code provides an interpretable and editable representation for 3D content. Existing methods generate shape programs, procedural models, CAD code, or Blender scripts from language and visual inputs ~\cite{jones2020shapeassembly,sun20253d,du2412blenderllm,hu2024scenecraft,lu2025ll3m,yin2026vision}. Related work studies part-level object representations and articulated-asset generation ~\cite{mo2019partnet,xiang2020sapien,lei2023nap,liu2024cage,chen2024urdformer,zhou2026articraft}. Most relevant to our setting, SceneCode generates part-aware object programs and compiles them into simulation assets with inferred revolute and prismatic joints ~\cite{wang2026scenecode}. Code-as-Room represents complete indoor scenes, including layouts, geometry, materials, and lighting as executable Blender code ~\cite{yang2026code}. Nevertheless, prior code-based systems primarily represent geometry, layout, appearance, or object-level kinematics. We instead generate functionally interactive scene programs that jointly encode articulated mechanisms, interaction states, and cross-entity behaviors such as switches controlling lights from user intent, making scenes both physically actionable and behaviorally coherent.

\section{Method}
\label{sec:method}

\subsection{Overview}
\label{sec:overview}

Given a text description of a room, \method produces a
fully instantiated, \emph{interactive} and \emph{physically executable} 3D scene:
a set of simulation-ready articulated assets, a metric layout that places each asset with a pose,
and an interaction script that specifies which controllers actuate which targets and with what effect. The output can be loaded directly into an Omniverse-based physics simulator such as NVIDIA Isaac Sim (or the OmniGibson~\cite{li2023behavior} environment built on it), where a robot can flip a switch and see the lamp turn on, turn a program dial and see the display change, or press a microwave button and see the panel light up. Unlike prior scene generators that stop at a static arrangement of
rigid props, our target is a scene whose every declared interaction can actually be triggered in simulation.

\subsection{\method Agentic System}

Figure~\ref{fig_framework} shows the overall architecture of \method, which
consists of three stages:

\textbf{(1) Scene Planning. } Given the room description, we first plan the
architectural shell of the room, including its shape, floor, and wall layout.
A usage-driven reasoning chain then expands the explicitly named objects
into a complete, operable scene state $S=(O,F,G_x)$, comprising an object set
$O$, a support forest $F$ recording, for each object, which object it rests on and on which supporting surface, and an interaction graph $G_x$ specifying how each object is operated. The chain
treats each major floor-standing anchor as a task centre, radially admits only
the companion objects that the task necessitates (growing the object set $O$
and the support forest $F$), and then \emph{exhaustively} induces, for every
object, its interaction edges $e=(s,t,m,\phi)$, which link controllers,
targets, and named parts under a closed method vocabulary, so that the scene
holds only what its use needs but exposes everything that use can touch.

\textbf{(2) Simulation-ready Asset Generation.} 
We use Articraft~\cite{zhou2026articraft} as the backend for articulated-object code generation. To preserve the LLM's spatial understanding acquired during modeling, we additionally extract part-level annotations and their 3D positions in the asset coordinate frame:
\begin{equation}
  A = \{(\ell_i, \mathbf{p}_i)\}_{i=1}^{N},
\end{equation}
where $\ell_i$ denotes the semantic label of part $i$, and $\mathbf{p}_i \in \mathbb{R}^3$ denotes its local position.

To support physics-based interaction, each asset is compiled into addressable links with collision geometry and dynamic properties. Hinged and sliding parts are converted into revolute and prismatic joints with inferred origins, axes, and limits. For each link $\ell$, we estimate
\begin{equation}
  \pi(\ell)=
  \big(m_\ell,c_\ell,I_\ell,\mu_\ell,\mathcal{C}_\ell\big),
  \label{eq:phys}
\end{equation}
where $m_\ell$, $c_\ell$, $I_\ell$, $\mu_\ell$, and $\mathcal{C}_\ell$ denote its mass, center of mass, inertia, friction, and simplified convex collision geometry, respectively. 

To support interactions beyond object-level articulation, we compile each edge $e=(s,t,m,\phi)$ in the interaction graph $G_x$ into a trigger–condition–effect rule over object state. The trigger is derived from method $m$, the conditions from the interaction's prerequisites, and the effect applies $\phi$ as the resulting  object-state or visual-property update. The resulting assets, poses, and interaction runtime are directly loaded into OmniGibson~\cite{li2023behavior} for scripted validation and embodied-agent interaction. 

To support coherent scene appearance, we use FLUX.2~\cite{flux-2-2025} to generate tileable albedo maps for architectural surfaces and planar decorations, which are applied using closed-form UV projections. Objects retain their part-level semantic colors, while PBR metalness and roughness are assigned according to material class. This process produces a consistent appearance without explicit UV unwrapping or per-object texturing.

\textbf{(3) Layout Generation.} A symbolic layout program
expressed in the user's reference frame (support relations, user-viewpoint spatial relations, and usage-guided orientations for manipulands) is grounded by a geometric solver into collision-free poses. We then introduce a vision-in-the-loop critic that refines the provisional scene according to rendered observations and the upstream layout guidance. The critic evaluates visual realism, functional consistency, spatial organization, and completeness, and invokes scene-editing to adjust object placement, orientation, and inter-object relations. Each revision is validated by physics-simulation tools, which resolves geometric conflicts and rejects modifications that introduce collisions, instability, or other physical violations. This iterative process preserves the intended high-level organization while improving the visual and physical plausibility of the final layout.

\subsection{Usage-Driven Object Reasoning}
\label{sec:reason}

\paragraph{}
Formally, given a room description $d$ and the bare set $O_P$ of objects it
explicitly names, the objective of scene reasoning is to expand this seed into a
complete, \emph{operable} scene state
\begin{equation}
  S \;=\; \big(\,O,\; F,\; G_x\,\big),
  \label{eq:state}
\end{equation}
comprising an object set $O\supseteq O_P$, a \emph{support forest} $F$ recording,  for each object, which object it rests on and on which supporting surface, and an \emph{interaction graph} $G_x$
specifying how each object is operated. We implement this expansion as a
\textbf{usage-driven reasoning process}. Specifically, we formulate reasoning as a \emph{hierarchical expansion} that
descends the room's functional structure rather than flattening it into a prompt
transcription. 

\begin{algorithm}[t]
\caption{Usage-Driven Object Reasoning}
\label{alg:reason}
\begin{algorithmic}[1]
\Require room description $d$
\Statex \textbf{Zoned completion (one pass).}
\State $(Z, O, F) \gets \mathcal{R}(d)$
  \Comment{zones, seed + inferred objects, support forest $F$}
\Statex \textbf{Phase 1 — Task-centric expansion (radial, per anchor).}
\ForAll{major floor anchor $a \in O$}
  \State $\tau_a \gets \mathcal{R}(a, d)$
    \Comment{name the human task $a$ centres}
  \State $C_a \gets \mathcal{R}(a, \tau_a, O)$
    \Comment{one-hop companions the task needs; necessity-gated top-$k$}
  \State $O \gets O \cup C_a;\quad F \gets F \cup \{(c \to a) : c \in C_a\}$
\EndFor
\Statex \textbf{Phase 2 — Interaction induction (exhaustive, per object).}
\ForAll{object $o \in O$}
  \State $G_x \gets G_x \cup \mathcal{R}(o)$
    \Comment{every affordance a user actuates, uncurated}
\EndFor
\State \Return $S = (O,\, F,\, G_x)$
\end{algorithmic}
\end{algorithm}

Prior methods typically enumerate objects sequentially according to a predefined spatial order or infer them from local visual observations. Consequently, the generated object composition lacks a systematic connection to how the room is intended to be used. Our process instead grounds every decision in usage mainly across
two phases (Algorithm~\ref{alg:reason}): each object is admitted because an activity requires it, and each interaction because it is a way a user operates the object that bears it. The result is a coherent \emph{functional hierarchy}, comprising anchors, the tasks they centre, the objects those tasks demand, and the affordances those objects expose, rather than an unstructured collection of objects.

\paragraph{Task-Centric Expansion.}
The reasoner treats each major floor anchor as a \emph{task centre}: it first identifies the activity the anchor supports, and only then determines the objects that activity requires within a user's reach. Expansion is \emph{radial, not chained}: every candidate must connect to the anchor's task within a single hop, preventing drift through long associative chains. It is also \emph{necessity-gated}: candidates are scored by how essential they are to the task, and only the most essential are admitted, with the rest held as an optional pool. Admitted objects extend both $O$ and the support forest $F$, each bound to the anchor whose use justifies it.

\paragraph{Interaction Induction.}
For each object the reasoner induces its interaction edges, $e \;=\; \big(s,\; t,\; m,\; \phi\big)\in G_x$, each linking a source $s$ to a target $t$, which may be the object itself, another object, or a named part $o\#\mathrm{part}$, under an interaction method $m$ from a closed vocabulary $\mathbb{M}$, with $\phi$ the resulting change of state. In contrast to the necessity-gated completion, induction is \emph{exhaustive} and \emph{functional}: every affordance a user could actuate becomes a realizable edge, derived from how the object is used rather than its category. The part names referenced by edges form a naming contract carried into asset generation.

\subsection{Code-Based Multi-part Interaction}
For each object, the reasoner induces its interaction edges, $e \;=\; \big(s,\; t,\; m,\; \phi\big)\in G_x$ , a code agent then compiles each edge into executable behavior. Rather than emitting isolated articulation parameters, the agent lowers every edge into a structured specification comprising the target's mutable state together with a trigger, a set of conditions, and a set of effects that transition it, and pairs it with a \emph{grounding} that resolves each abstract role in the edge to a concrete, addressable entity in the simulated asset: the trigger is the actuation signal that initiates the interaction (a joint reading or a contact event), the conditions are the prerequisites under which it is valid, and the effects specify how the kinematic, physical, or visual state of the target is updated. These specifications are executed by a single generic runtime shared across all assets. Since edges reference named parts rather than fixed geometric primitives, the compiled behavior is decoupled from any specific asset implementation: the part names carried by each edge form a naming contract into asset generation, so that every control and feedback element is later realized as an addressable link to which the runtime binds, and where an edge requires a controller the scene lacks, that controller is admitted into $O$. Expressing interactions as structured, runtime-executed code rather than isolated articulation parameters makes behaviors inspectable, editable, and composable, and lets coordinated behavior across multiple parts or objects emerge from the joint execution of individual edges.



\subsection{Usage-guided Orientation for Small Manipulands}
\label{sec:orient}

Manipuland orientation is difficult to infer from visual observations alone. Moreover, the notion of \texttt{front} is often ambiguous for manipulands, as their appropriate orientation is determined by intended use rather than visual geometry alone. 

Different from prior methods, we first introduce the annotation information during code-based modeling of object. Our key observation is that code-based asset generation already exposes stronger signals than downstream pixels. We fuse the annotation map $A$ with a small set of rendered views of the asset. The annotations contribute precise semantic structure; the images contribute holistic shape and appearance cues. A VLM reasons over this fused representation and predicts an intrinsic orientation frame $(\mathbf{f},
\mathbf{u})$.

To obtain a user-centric initialization of the front direction, we ask the VLM to predict two additional anchors in the normalized bounding-box frame: a hand anchor $\mathbf{h}$, where
the user would contact or operate the object, and an arm anchor $\mathbf{a}$,
indicating the approach side of the body. Their displacement
\begin{equation}
  \mathbf{d}_{\mathrm{use}} = \mathbf{h} - \mathbf{a}
\end{equation}
provides a geometric cue for how the object should face in use. We align this vector to the nearest principal box axis orthogonal to the estimated up axis, producing the final use-facing direction $\mathbf{f}_{\mathrm{use}}$, which is used to initialize the object's pose during placement.

\section{Experiments}

\begin{figure*}[h]
  \centering
   \includegraphics[width=1\linewidth]{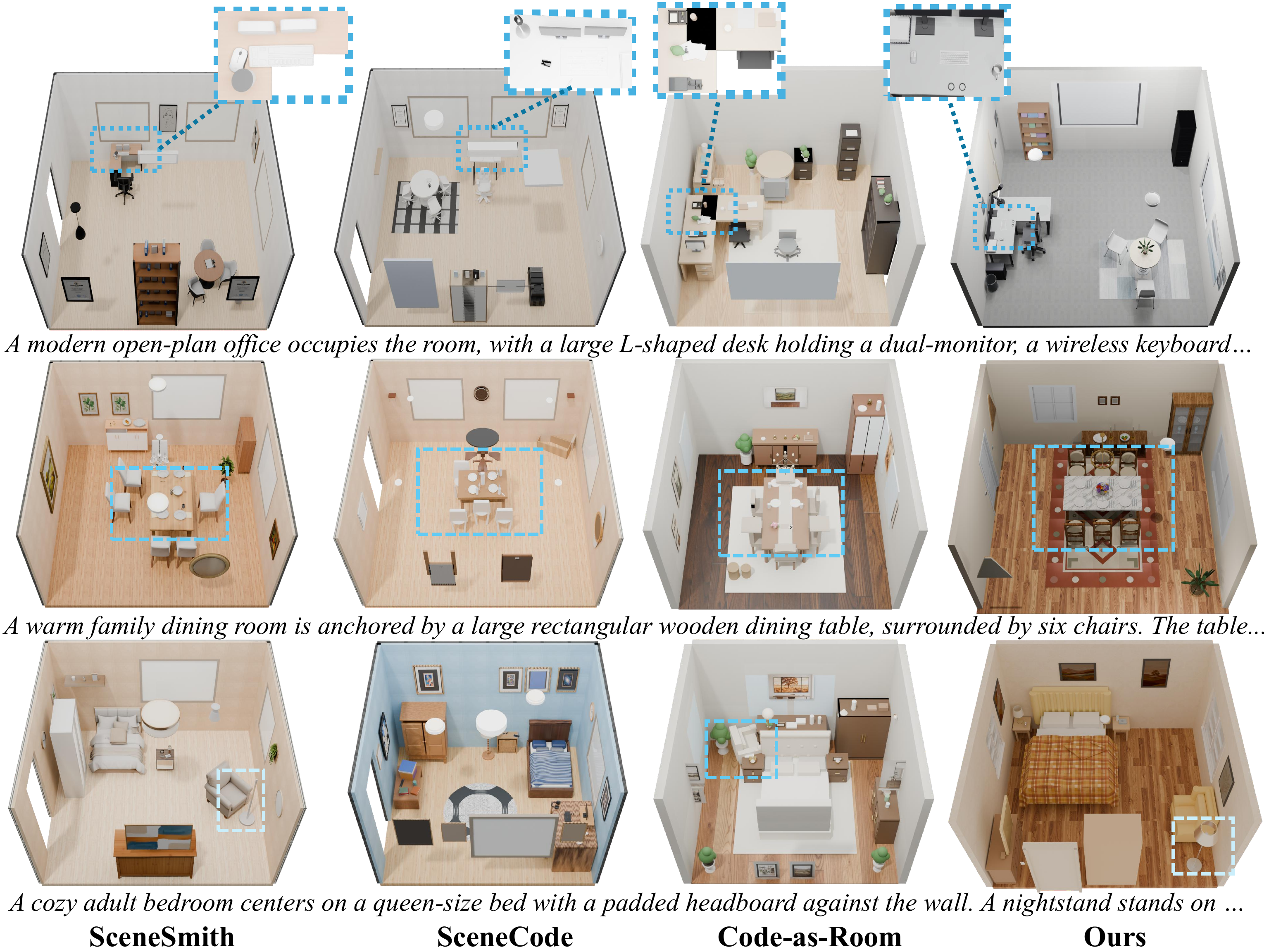}

   \caption{
   Qualitative comparison of the four methods. \method produces layouts with better plausibility than the baselines, and achieves highly realistic visual quality even though it relies entirely on code, without any 3D generation or asset retrieval.
}
   \label{fig_overall}
\end{figure*}

\begin{table*}[t]
  \centering

  \resizebox{\textwidth}{!}{%
  \begin{tabular}{lcccccccccc}
  \toprule
  Method & Collision-free $\uparrow$ & Navig. $\uparrow$ & In-bound $\uparrow$ & Opening-clear $\uparrow$ & Support
  $\uparrow$ & Access. $\uparrow$ & ObjCount $\uparrow$ & ObjAttr $\uparrow$ & OO-Rel $\uparrow$ & OA-Rel
  $\uparrow$ \\
  \midrule

  SceneCode & 91.2 & \underline{99.0} & 81.2 & \underline{96.4} & 26.9 & 26.1 & 38.1 & 30.7 & 16.1 & \textbf{55.6} \\
  Code-as-Room & 45.5 & 89.5 & 73.3 & 47.2 & 51.9 & \underline{63.1} & 37.0 & \underline{56.7} & 20.5 & 22.2 \\
    SceneSmith & \underline{95.0} & \textbf{99.6} & \underline{99.7} & \textbf{98.0} & \textbf{58.4} & 39.8 & \underline{48.8} & 42.5 & \textbf{27.0} & \underline{42.2} \\
  \midrule
  \textbf{Ours} & \textbf{95.9} & 97.3 & \textbf{99.7} & 95.7 & \underline{53.6} &
  \textbf{65.6} & \textbf{57.0} & \textbf{64.6} & \underline{22.4} & 40.6 \\
  \bottomrule
  \end{tabular}%
  }
  \caption{SceneEval comparison across four scene-generation methods
  (mean over 28 scenes). Higher is better ($\uparrow$) for all
  metrics. The best result in each column is shown in \textbf{bold},
  and the second-best result is \underline{underlined}.}
  \label{tab_sceneeval_full}
\end{table*}

\paragraph{Baselines.}
Code-based full-scene generation is a recently emerging task with only a few directly comparable methods. We compare our approach with SceneCode~\cite{wang2026scenecode} and Code-as-Room~\cite{yang2026code}, two representative systems that generate complete indoor scenes using executable code. SceneCode supports part-aware articulated objects and simulation-ready asset compilation, whereas Code-as-Room synthesizes room-scale geometry, materials, and lighting as Blender programs. We additionally compare with SceneSmith~\cite{pfaff2026scenesmith}, a retrieval-based agentic framework that serves as the scene-generation backbone of SceneCode and has demonstrated substantial relevance to embodied AI.

\paragraph{Input text descriptions.}
To comprehensively evaluate the capabilities of each method, we curate 28 room-level prompts covering diverse room categories and functional requirements. The prompts are organized into three difficulty levels according to their descriptive complexity: short and concise descriptions, medium-length descriptions with additional spatial and functional constraints, and long, detailed descriptions specifying complex scene composition and interactions. The complete prompt set is provided in the supplementary material.

\paragraph{Automatic evaluation.} We adopt the scene-level
  metrics from SceneEval~\cite{tam2026sceneeval}: Collision-free (fraction of objects
  free of inter-object collision), Navig. (navigable floor-area
  ratio after eroding object footprints), In-bound (fraction of
  objects lying within the room bounds), Opening-clear (fraction of
  doors and windows left unobstructed), Support (fraction of
  objects resting on a valid supporting surface rather than
  floating or clipping), Access. (reachability of each object's
  functional side), ObjCount (satisfaction of object-count
  requirements), ObjAttr (satisfaction of object-attribute
  requirements), OO-Rel (satisfaction of object--object
  spatial-relationship requirements), and OA-Rel (satisfaction of
  object--architecture relationship requirements). All metrics are
  reported as percentages and higher is better.

\begin{table}[h]
\centering

\setlength{\tabcolsep}{5.5pt}
\renewcommand{\arraystretch}{1.08}
\begin{tabular}{lcccc}
\toprule
Method & IO$\uparrow$ & AO$\uparrow$ & CHI$\uparrow$ & MPC$\uparrow$ \\
\midrule
Code-as-Room & -- & -- & -- & -- \\
SceneSmith   & 2.4 & 2.4 & -- & -- \\
SceneCode    & 4.4 & 4.4 & -- & -- \\
\hline
Ours         & \textbf{16.4} & \textbf{13.2} & \textbf{6.4} & \textbf{3.6} \\
\bottomrule
\end{tabular}
\vspace{2pt}
\begin{minipage}{0.98\linewidth}
\footnotesize

\end{minipage}
\caption{Interactive objects and behaviors in generated scenes. '--' indicates that the method does not provide the corresponding attribute.}
\label{inter_nums}
\end{table}

Furthermore, we evaluate interactive content using four metrics: Interactive Objects (IO), objects supporting executable actions or state changes; Articulated Objects (AO), objects with revolute or prismatic joints; Complex Non-Hinge Interactions (CHI), functional relations beyond independent articulation; and Multi-Part Coordination (MPC), coordinated interaction groups involving multiple parts or objects. IO and AO are counted per object, CHI per functional relation, and MPC per coordination group; thus, these metrics are complementary rather than additive.

\subsection{Overall Comparisons}
Quantitative results are shown in Table \ref{tab_sceneeval_full}, while Figure \ref{fig_overall} shows the visual quality. SceneCode lacks holistic scene-level planning and instead generates code through a sequential, step-by-step VLM process. Consequently, local code-generation errors may accumulate and become amplified during scene assembly, reducing overall quality. Code-as-Room produces the lowest visual quality in our comparison, as it provides limited physical guarantees and relies on a relatively coarse code-generation pipeline, resulting in approximate scene geometry and non-interactive objects. SceneSmith generates visually appealing scenes but similarly adopts a sequential planning strategy and constructs scenes primarily through asset retrieval or image-to-3D generation. Its outputs therefore do not inherit the interpretability and editability of fully code-based representations, and most objects remain non-interactive. In contrast, our method achieves the highest overall quality among the evaluated code-based approaches, producing detailed and complete scenes. Its  user-centered composition improves asset completeness and layout plausibility, while supporting substantially richer interactive behaviors. Table~\ref{inter_nums} reports the number of interactive objects. Our method generates more interactive objects than the baselines and, beyond articulated objects, supports a diverse range of complex interaction types that are not available in the other methods.

\begin{figure}[t]
  \centering
   \includegraphics[width=1\linewidth]{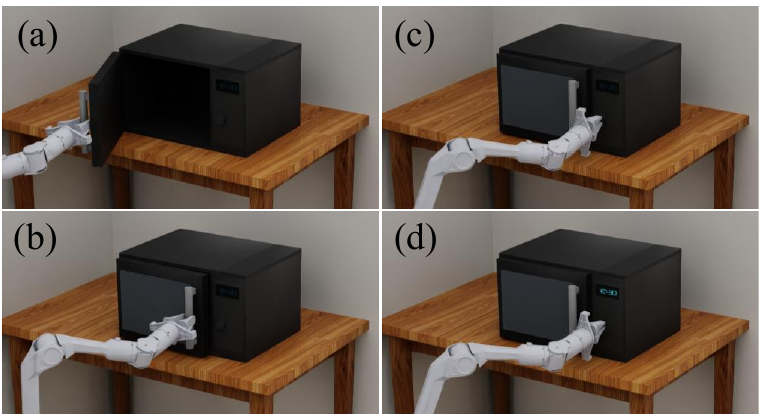}

   \caption{Simulation of generated interactions. (a)(b) The gripper grasps the handle and closes the revolute door. (c) The arm approaches the button while the display remains off. (d) Pressing the button lights up the display, demonstrating a causal interaction.}
   \label{fig_fobot}
\end{figure}

\subsection{Ablation Study}

\begin{figure}[t]
  \centering
   \includegraphics[width=1\linewidth]{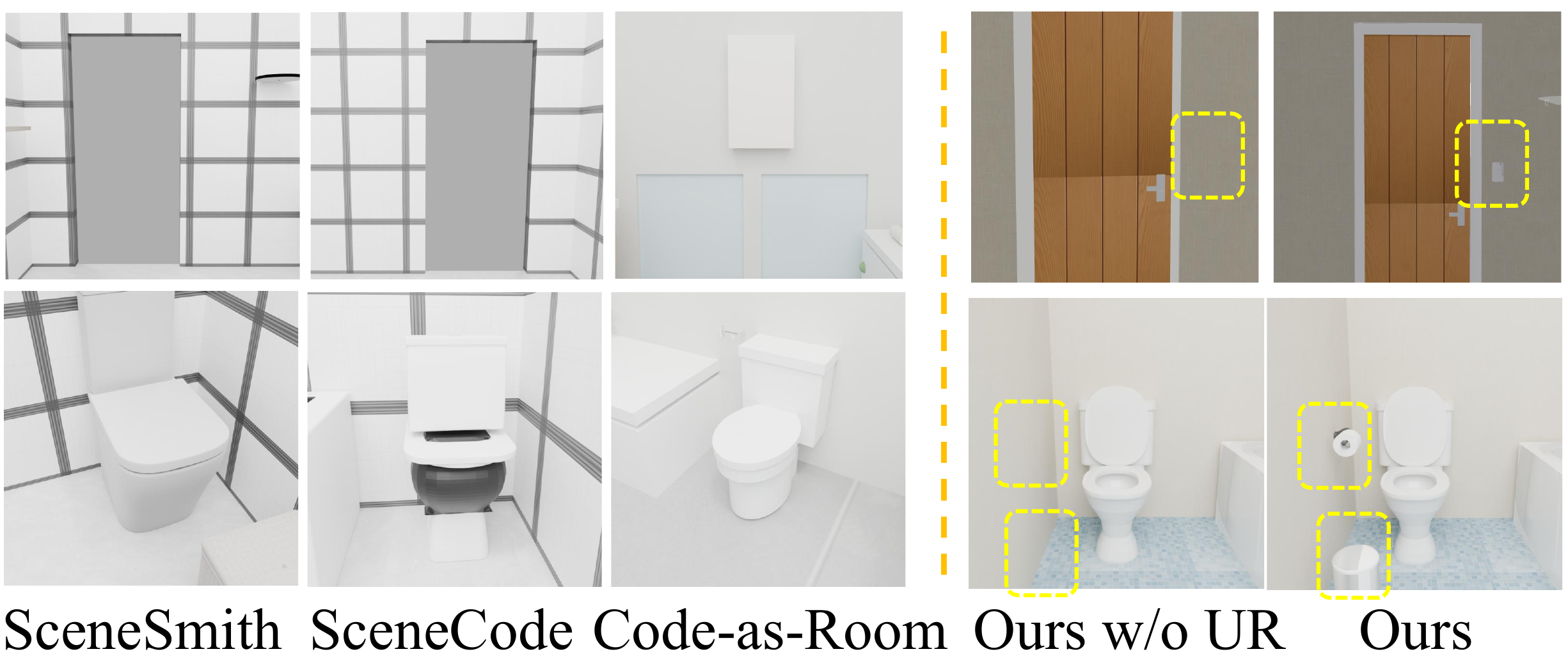}

   \caption{
   Qualitative comparison with and without usage-driven object reasoning (UR in the figure).
}
   \label{fig_user_reasoning}
\end{figure}

\paragraph{Ablation study of usage-driven object reasoning}
As shown in Figure~\ref{fig_user_reasoning}, the usage-driven object reasoning chain produces a richer and more functionally plausible object composition, including contextually necessary objects and accessible interaction controls. These objects are absent when the usage-driven reasoning chain is removed. SceneCode, SceneSmith, and Code-as-Room also fail to generate them, revealing a limitation of prior methods that construct scenes through sequential object-by-object planning without explicitly reasoning about intended use.

\begin{figure}[t]
  \centering
   \includegraphics[width=1\linewidth]{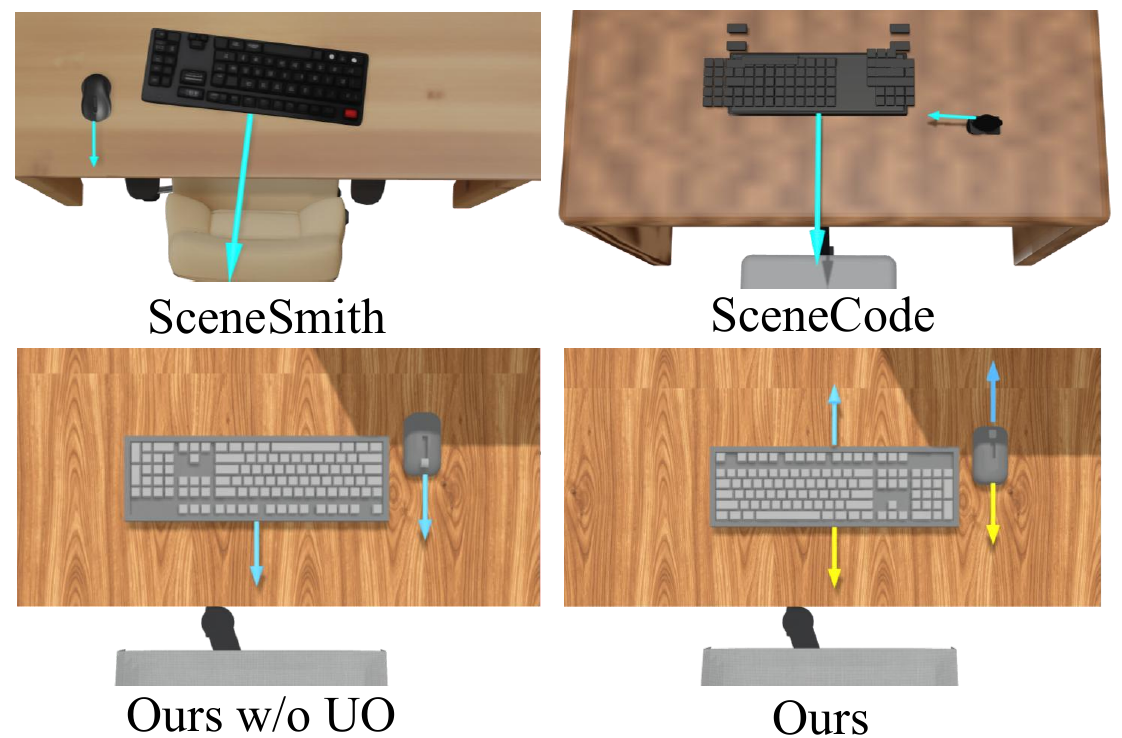}

   \caption{
   Comparison of manipuland placement with and without usage-guided orientation (UO in the figure).
}
   \label{fig_orientation}
\end{figure}

\paragraph{Ablation study of usage-guided orientation}
Figure~\ref{fig_orientation} demonstrates the effectiveness of usage-guided orientation. The blue arrows indicate the front directions predicted by the VLM, while the yellow arrows denote the use-facing direction inferred by our agent. In all four cases, object placement is performed directly by the corresponding agentic system. Our method preserves detailed part-level annotations from the initial modeling stage, including individual keyboard keys as well as the mouse buttons and scroll wheel. This avoids requiring the VLM to recover critical orientation cues from visually ambiguous rendered views. The inferred use-facing direction further guides object placement, enabling our method to be the only one that correctly orients and arranges both the keyboard and mouse.

\subsection{Robot interaction}
As shown in Figure~\ref{fig_fobot}, our scene can be imported into Isaac Sim for robot manipulation. Movable parts of objects remain independent links with compiled joints. 
Beyond geometry, each asset carries a generated interaction manifest, including actuator joints, responder materials, and trigger rules, which is executed by a generic runtime that closes the loop from actuation to feedback. 

\section{Conclusion}
We presented \method, a usage-driven framework that generates 3D scenes represented entirely as code. By deriving scene content from intended activities, enabling multi-part interaction through structured behavior code, and assigning usage-guided orientations from modeling-time annotations, \method goes beyond visual construction and object-level articulation to model how a scene is actually used. Experiments show that it produces plausible, realistic, and executable interactive scenes for embodied AI.
\bibliography{aaai2027}


\newpage
\clearpage

\section{Appendix}

This appendix provides additional technical details, results, and analyses of \method. The contents are organized as follows:
\begin{itemize}
\item A ~\ref{sec:prompts}Evaluation Prompt Suite
\item B ~\ref{res} Additional Results
\item C ~\ref{sec:metrics}  Evaluation Metrics
\item D ~\ref{interact_obj} Interactive Object Demonstrations
\item E ~\ref{edit} Editability of Code-Represented 3D Assets
\item F ~\ref{cost} Computational Cost Statistics
\item G ~\ref{lim} Limitation

\end{itemize}



\section{A. Evaluation Prompt Suite}
\label{sec:prompts}

Our main evaluation uses a suite of \textbf{28 natural-language scene
prompts}, reproduced verbatim below. The suite is organized into three
difficulty tiers---16 \emph{easy}, 7 \emph{medium}, and 5 \emph{hard}
prompts---and spans 23 distinct room types. This section describes how the
difficulty gradient and the room-type coverage were designed, and then lists
every prompt.

\paragraph{Difficulty gradient.}
Difficulty is controlled along four axes that stress progressively deeper
stages of a scene-generation pipeline:

\begin{itemize}
  \item \textbf{Mentioned-object count.} Easy prompts name 2--4 anchor
  objects; medium prompts name 4--8; hard prompts describe 15--20 entities,
  including small props (an alarm clock, a glass of water, candlesticks) in
  addition to furniture.
  \item \textbf{Explicit cardinality.} Medium and hard prompts introduce
  counts that must be satisfied exactly (\emph{six chairs}, \emph{three bar
  stools}, \emph{two rows of theater seats}), which one-shot layout methods
  frequently violate.
  \item \textbf{Relational constraints.} Easy prompts are flat object lists.
  Medium prompts add a single spatial relation (\emph{an island in the
  middle}, \emph{a sideboard against the wall}). Hard prompts are
  paragraph-length and encode nested, chained relations (\emph{a nightstand
  on each side of the bed, the left one holding \dots}, \emph{an armchair in
  the corner near the window with a floor lamp beside it}), requiring the
  layout solver to compose many constraints consistently.
  \item \textbf{Surface and wall clutter.} On-surface props and wall-mounted elements appear from the medium tier. The hard tier 
  additionally requires dense surface clutter and wall decoration, exercising the on-surface and wall-mounting stages at full load.
\end{itemize}

The five hard prompts are intentionally written as flowing descriptive
paragraphs rather than lists, so that object completion (recovering implied
items), relation extraction, and counting must all be inferred from natural
language rather than read off a template.

\paragraph{Room-type diversity.}
The 28 prompts cover 23 room types drawn from five functional categories:
(i)~\emph{core residential} rooms (bedroom, bathroom, kitchen, living room,
dining room, office, gym); (ii)~\emph{utility} spaces (laundry room, pantry,
mudroom, garage workshop); (iii)~\emph{work and creative} spaces (art
studio, music room, sewing room, photography studio, home library);
(iv)~\emph{leisure} spaces (home bar, wine cellar, home theater, greenhouse
sunroom, children's playroom); and (v)~\emph{commercial} interiors (dental
exam room, barbershop). This mix tests generalization across both
object vocabularies (a dental chair shares little with a drum kit) and
layout conventions (row seating in a theater vs.\ radial seating around a
dining table).

Crucially, five room types---\emph{office, bedroom, bathroom, dining room,
and gym}---appear in \textbf{both} the easy tier and the hard tier. These
form controlled within-type pairs (e.g., \texttt{bathroom1} vs.\
\texttt{bathroom2}) in which the room category and its core furniture are
held fixed while the specification complexity is scaled up, allowing
difficulty effects to be measured independently of room-type effects. Scene
identifiers below follow this convention: a trailing \texttt{2} denotes the
hard-tier member of a duplicated type.

\subsection{Easy prompts (16)}

Table~\ref{tab:prompts_easy} lists the 16 easy-tier prompts verbatim.

\begin{table*}[t]
  \centering
  \small
  
  \begin{tabular}{@{}llp{0.62\linewidth}@{}}
    \toprule
    ID & Room type & Prompt (verbatim) \\
    \midrule
    \texttt{laundry\_room} & laundry room &
A laundry room with a washing machine and a dryer. \\
\texttt{bathroom1} & bathroom &
A bathroom with a vanity, a bathtub, and a toilet. \\
\texttt{office1} & home office &
A home office with a desk, an office chair, and a bookshelf. \\
\texttt{gym1} & home gym &
A home gym with a treadmill and an exercise bike, and a wall mirror. \\
\texttt{library} & home library &
A home library with a bookshelf, a reading armchair, and a floor lamp. \\
\texttt{art\_studio} & art studio &
An art studio with an easel, a stool, and a worktable. \\
\texttt{music\_room} & music room &
A music room with a piano and a piano bench, and a drum kit. \\
\texttt{home\_bar} & home bar &
A home bar with a bar counter and three bar stools, and a wine rack. \\
\texttt{sunroom} & greenhouse sunroom &
A greenhouse sunroom with a potting bench, plant stands, and a lounge chair. \\
\texttt{wine\_cellar} & wine cellar &
A wine cellar with wine racks, a tasting table, and four chairs. \\
\texttt{photo\_studio} & photography studio &
A photography studio with a backdrop, two studio lights, and a camera on a tripod. \\
\texttt{sewing\_room} & sewing room &
A sewing room with a worktable, a sewing machine, and a fabric shelf. \\
\texttt{mudroom} & mudroom &
A mudroom with a bench, wall hooks, and a shoe cabinet. \\
\texttt{pantry} & pantry &
A pantry with tall shelving units, a wire rack, and a narrow counter. \\
\texttt{dental\_office} & dental exam room &
A dental exam room with a dental chair, an overhead light, and a counter with a sink. \\
\texttt{barbershop} & barbershop &
A barbershop with three barber chairs, a wall mirror, and a row of waiting chairs. \\
    \bottomrule
  \end{tabular}
  \caption{The 16 easy-tier evaluation prompts (verbatim).}
  \label{tab:prompts_easy}
\end{table*}

\subsection{Medium prompts (7)}

Table~\ref{tab:prompts_medium} lists the 7 medium-tier prompts verbatim.

\begin{table*}[t]
    \centering
    \small

\begin{tabular}{@{}llp{0.62\linewidth}@{}}
      \toprule
      ID & Room type & Prompt (verbatim) \\
      \midrule
      \texttt{kitchen} & kitchen &
  A kitchen with a counter along the main wall holding a microwave and a knife block, a refrigerator at one end of the counter, and a built-in oven with
  a stovetop, with an island in the middle with two bar stools tucked underneath. \\
  \texttt{bedroom1} & bedroom &
  A bedroom with a queen bed against the main wall, a nightstand with a table lamp on each side of the bed, a tall wardrobe with sliding doors, a low
  dresser with a wide mirror above it, and a cushioned bench at the foot of the bed. \\
\texttt{living\_room} & living room &
  A living room with a three-seat sofa facing a wall-mounted TV, a coffee table with a stack of magazines on a wide area rug, a media console beneath the
  TV flanked by two tall floor speakers, and a bookshelf along the side wall. \\
\texttt{dining\_room1} & dining room &
  A dining room with a long rectangular table surrounded by six chairs and set with tableware, a sideboard against the wall holding serving dishes, and a
  pendant chandelier above the table. \\
\texttt{garage\_workshop} & garage workshop &
  A garage workshop with a heavy workbench along the main wall fitted with a bench vise and a toolbox, a tall pegboard above the bench holding hanging
  tools, a rolling tool chest with drawers, and a metal shelving unit against the side wall. \\
\texttt{playroom} & children's playroom &
  A children's playroom with a low activity table with four small chairs, a tall toy shelf filled with bins of toys, a cozy reading corner with a bean
  bag chair and a floor lamp, and a play tent near the window. \\
\texttt{home\_theater} & home theater &
  A home theater with a large projection screen on the main wall faced by two rows of theater seats, a projector mounted on the ceiling, a low media
  console beneath the screen flanked by two tall floor speakers, and a subwoofer in the corner. \\
      \bottomrule
    \end{tabular}
    \caption{The 7 medium-tier evaluation prompts (verbatim).}
    \label{tab:prompts_medium}
\end{table*}

\subsection{Hard prompts (5)}

The five hard prompts are paragraph-length descriptions of the five room
types that also appear in the easy tier, enabling the controlled
within-type comparison described above.

\begin{description}
  \item[\texttt{office2} (open-plan office)]
  A modern open-plan office occupies the room, with a large L-shaped desk
  against the window holding a dual-monitor setup, a wireless keyboard and
  mouse, a desk lamp, and a stack of documents in a tray. An ergonomic mesh
  office chair sits in front of the desk. Along the opposite wall stands a
  tall bookshelf filled with binders and books, beside a black filing
  cabinet with several drawers. A small round meeting table with three
  chairs sits in one corner, with a potted plant nearby for greenery. A
  whiteboard hangs on the wall above the meeting table, and a wall clock is
  mounted near the door. A patterned area rug anchors the seating area, and
  framed certificates decorate the walls.

  \item[\texttt{bedroom2} (adult bedroom)]
  A cozy adult bedroom centers on a queen-size bed with a padded headboard
  against the main wall, dressed with pillows and a folded blanket. A
  nightstand stands on each side of the bed, the left one holding a table
  lamp and an alarm clock, the right one holding a book and a glass of
  water. Opposite the bed, a wide wooden dresser with multiple drawers
  supports a framed mirror and a small jewelry box. In the corner near the
  window sits a comfortable armchair with a floor lamp beside it for
  reading. A tall wardrobe provides storage along the side wall, and a soft
  rectangular rug covers the floor beside the bed. Framed artwork and a few
  hanging plants decorate the walls.

  \item[\texttt{bathroom2} (modern bathroom)]
  A clean modern bathroom features a white pedestal sink with a
  wall-mounted mirror and a small shelf holding toiletries above it. Next
  to the sink stands a tall storage cabinet stocked with folded towels and
  bottles. A bathtub sits along the far wall with a shower head above it
  and a glass shelf nearby holding shampoo and soap. A toilet is placed in
  the corner with a small toilet-paper holder mounted beside it. A laundry
  hamper rests against the wall, and a bath mat lies on the floor in front
  of the tub. Wall hooks hold hanging robes and towels, and a small potted
  plant brightens the windowsill.

  \item[\texttt{dining\_room2} (family dining room)]
  A warm family dining room is anchored by a large rectangular wooden
  dining table surrounded by six matching upholstered chairs. The table is
  set with plates, glasses, cutlery, and a centerpiece vase of flowers.
  Against the wall stands a wide sideboard buffet holding a stack of
  serving dishes, a fruit bowl, and two candlesticks, with framed pictures
  hanging above it. A glass-front display cabinet in the corner showcases
  fine china and wine glasses. A pendant chandelier hangs above the center
  of the table, and a large area rug defines the dining space underneath. A
  potted plant sits near the window, and decorative artwork adorns the
  walls.

  \item[\texttt{gym2} (home gym)]
  A well-equipped home gym fills the room, with a treadmill positioned
  facing the mirror-covered wall and a stationary exercise bike beside it.
  A sturdy weight rack along one wall holds a row of dumbbells in
  increasing sizes, next to a flat workout bench. A yoga mat is rolled out
  on the floor with a foam roller and a stability ball nearby. A tall shelf
  stores folded towels, water bottles, and resistance bands. A wall-mounted
  clock and a motivational poster hang above the weight rack, and a small
  storage cabinet in the corner holds extra equipment. Rubber floor tiles
  cover the workout area for grip and cushioning.
\end{description}

\section{B. Additional Results}
\label{res}
\begin{figure*}[t]
  \centering
   \includegraphics[width=1\linewidth]{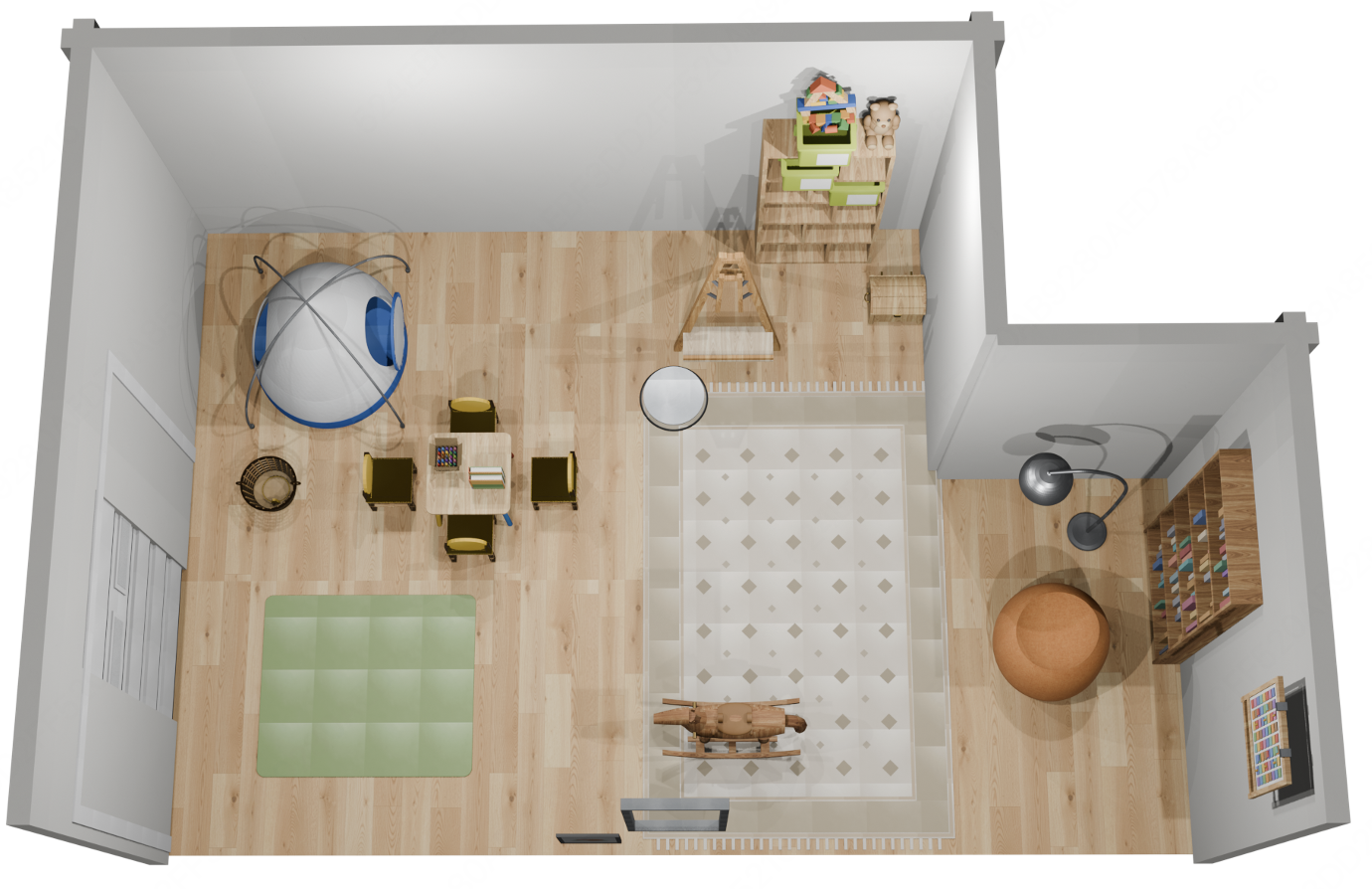}

   \caption{
   A children's playroom with a low activity table with four small chairs, a tall toy shelf filled with bins of toys, a cozy reading corner with a bean bag chair and a floor lamp, and a play tent near the window.
}
   \label{fig_play}
\end{figure*}

\begin{figure*}[t]
  \centering
   \includegraphics[width=1\linewidth]{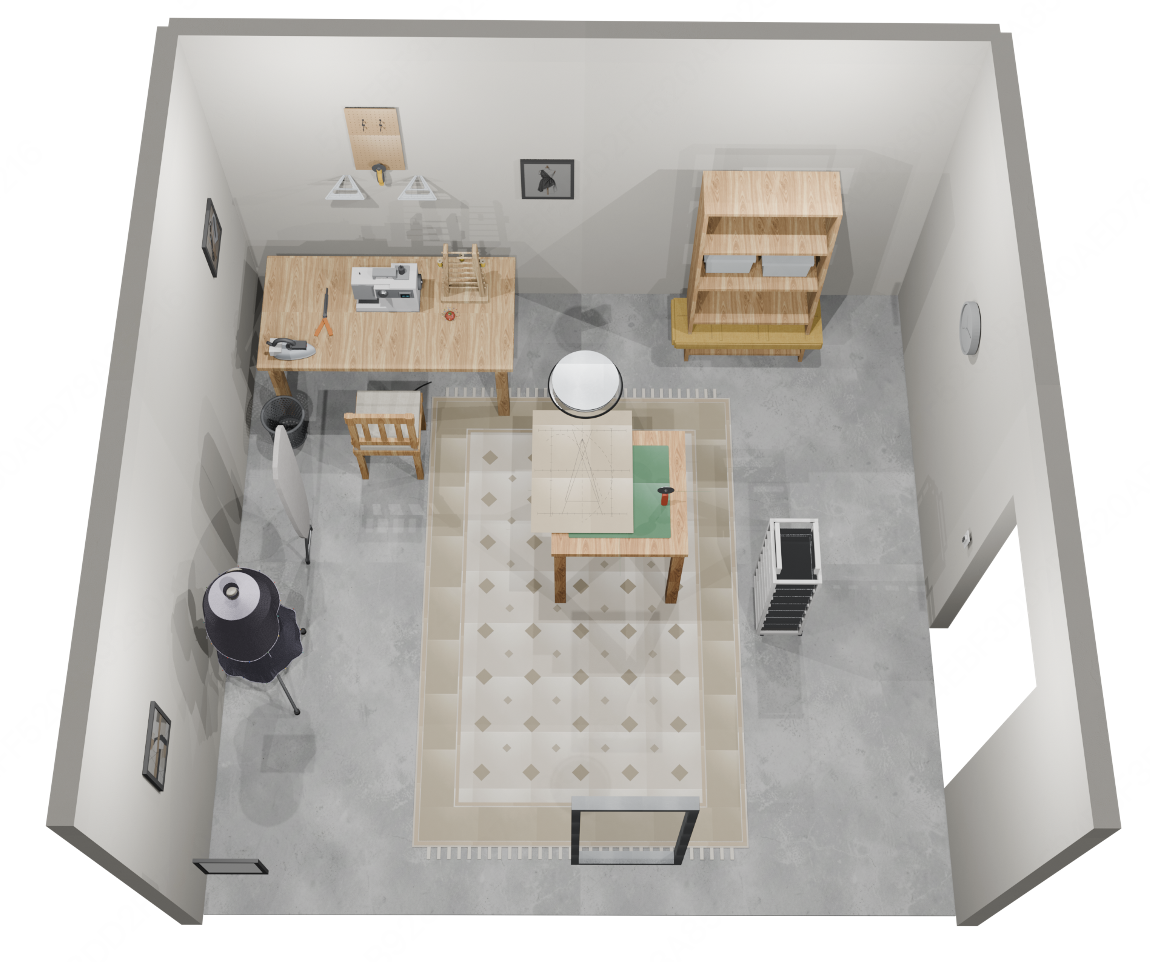}

   \caption{
   A sewing room with a worktable, a sewing machine, and a fabric shelf.
}
   \label{fig_sew}
\end{figure*}

\begin{figure*}[t]
  \centering
   \includegraphics[width=1\linewidth]{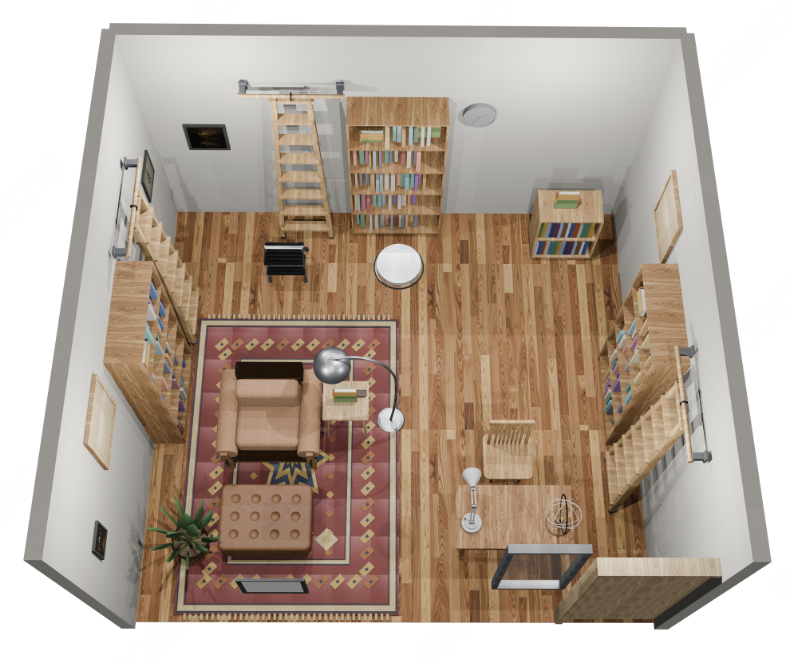}

   \caption{
   A home library with a bookshelf, a reading armchair, and a floor lamp.
}
   \label{fig_lib}
\end{figure*}

\begin{figure*}[t]
  \centering
   \includegraphics[width=1\linewidth]{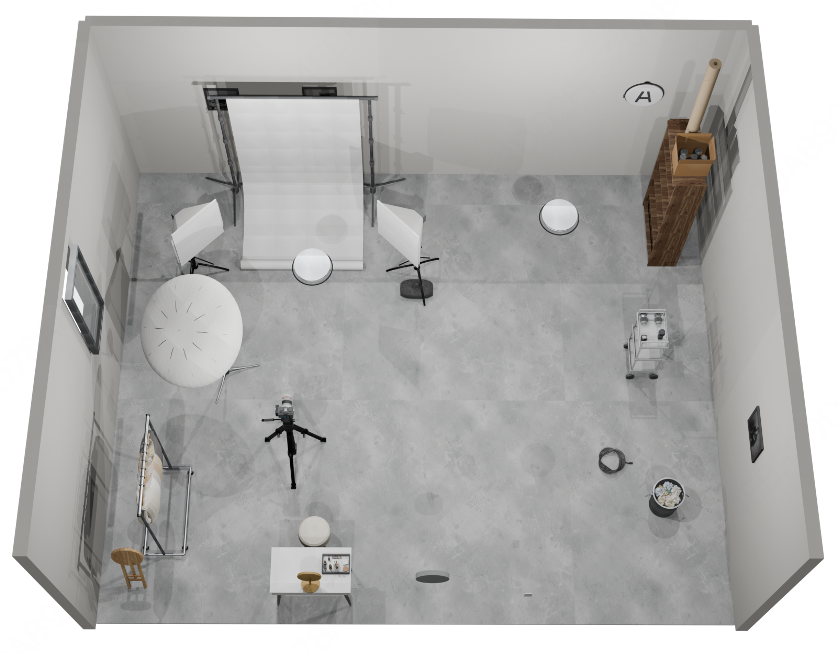}

   \caption{
   A photography studio with a backdrop, two studio lights, and a camera on a tripod.
}
   \label{fig_photo}
\end{figure*}

\begin{figure*}[t]
  \centering
   \includegraphics[width=1\linewidth]{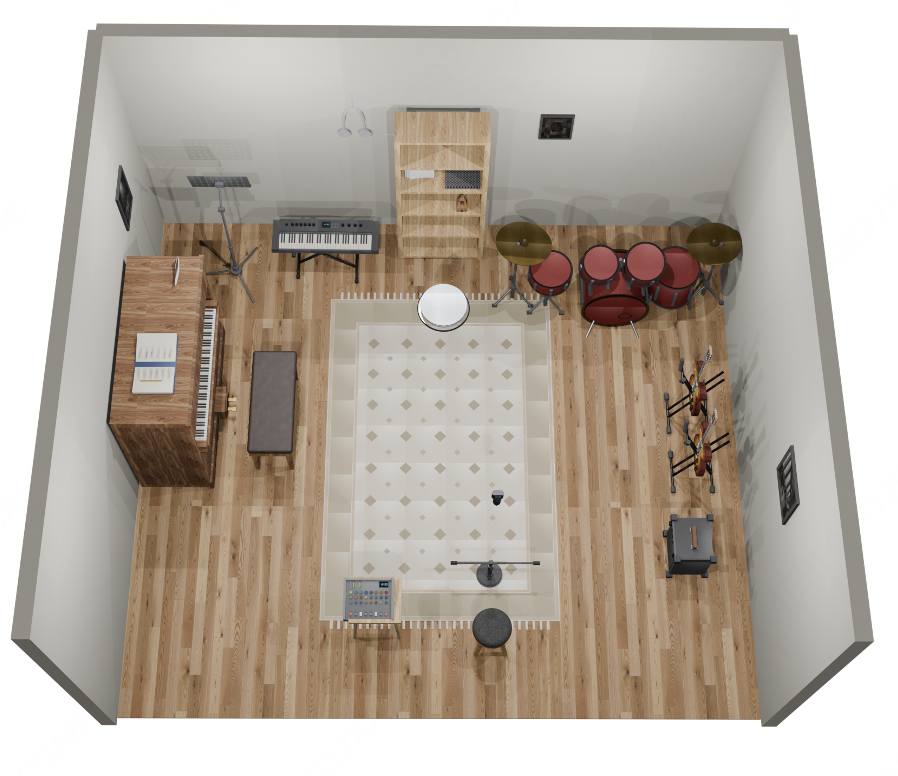}

   \caption{
   A music room with a piano and a piano bench, and a drum kit.
}
   \label{fig_music}
\end{figure*}

Figure~\ref{fig_play},~\ref{fig_sew},~\ref{fig_lib},~\ref{fig_photo} and~\ref{fig_music} additionally show several rooms of different types.

\section{C. Evaluation Metrics}
\label{sec:metrics}

We evaluate each generated scene on ten metrics from SceneEval, grouped into
three families (Table ~\ref{tab:metric_families}). Unless noted otherwise, every metric is reported as a fraction
in $[0,1]$ and \emph{higher is better}; throughout, $N$ denotes the number of
objects in the scene and $N_{\text{spec}}$ the number of annotated
requirements. Following the paper, we report the three count-style metrics
(Collision, Out-of-Bound, Opening-Clearance) as \emph{satisfaction ratios}
$1 - (\text{violations}/N)$ so that all ten columns share the ``higher is
better'' direction.

\subsection{Physical plausibility (geometry)}
\label{sec:metrics_phys}

\paragraph{Collision-free (COL).}
Two objects are in collision if their meshes interpenetrate. We use a two-stage
check: a pairwise penetration test, and—when an intersection is found—a
perturbation re-check that nudges the object and repeats the test to suppress
touching-contact false positives. Let $\mathbf{1}_{i}$ indicate that object $i$
interpenetrates at least one other object. The reported metric is the fraction
of non-colliding objects,
\begin{equation}
\mathrm{COL} \;=\; 1 - \frac{1}{N}\sum_{i=1}^{N} \mathbf{1}_{i}.
\end{equation}

\paragraph{Navigability (NAV).}
A top-down occupancy map of the floor is computed: the floor polygon is
rasterized, then \emph{eroded} by the robot footprint (width $w_{\text{robot}}
= 0.2\,$m), and object bounding boxes (footprints) are subtracted. We run
connected-component analysis on the free (walkable) pixels and report the area
of the largest connected free region normalized by the total free area,
\begin{equation}
\mathrm{NAV} \;=\; \frac{\bigl|\,\max\,\mathrm{CC}(\mathcal{W})\,\bigr|}{|\mathcal{W}|},
\end{equation}
where $\mathcal{W}$ is the walkable set after erosion and $\max\,\mathrm{CC}$
is its largest connected component. NAV $=1$ when the walkable space is a
single connected region.

\paragraph{In-bound (OOB).}
For each object we sample $k = \max(1000,\, 5000\, V_i)$ points on its surface
($V_i$ = oriented bounding-box volume) and cast a ray from each point in the
gravity direction $\mathbf{g}=(0,0,-1)$. A point is \emph{out of bound} if its
ray misses the room floor. With $r_i$ the in-bound point ratio, object $i$ is
out of bound when $r_i < 0.99$. The metric is the fraction of in-bound
objects,
\begin{equation}
r_i \;=\; 1 - \frac{\#\{\text{points of } i \text{ whose ray misses floor}\}}{k},
\end{equation}
\begin{equation}
    \mathrm{OOB} \;=\; \frac{1}{N}\sum_{i=1}^{N} \mathbf{1}[\,r_i \ge 0.99\,]
\end{equation}

\paragraph{Opening-clearance (OPC).}
For every door/window we extrude a clearance box into the room along the
opening's inward normal and test for intersecting objects. An opening is
\emph{clear} when no object interferes with its clearance volume. The metric is
the fraction of clear opening-direction checks,
\begin{equation}
\mathrm{OPC} \;=\; \frac{\#\{\text{opening checks with no interfering object}\}}
{\#\{\text{opening checks}\}}.
\end{equation}

\subsection{VLM-judged physical correctness}
\label{sec:metrics_vlm}

\paragraph{Support (SUP).}
A vision-language model first classifies each object's support type
(ground / wall / ceiling / on-object). Geometry then verifies the object is
supported: it is projected along the support's gravity vector and we require a
valid contact region whose centroid projection lies inside the contact-point
convex hull, within a distance threshold $\tau_{\text{sup}} = 0.01\,$m of the
supporting surface. The metric is the fraction of correctly supported objects,
\begin{equation}
\mathrm{SUP} \;=\; \frac{1}{N}\sum_{i=1}^{N} \mathbf{1}[\,d_i \le \tau_{\text{sup}}
\;\wedge\; \text{centroid}_i \in \mathrm{Hull}_i\,],
\end{equation}
where $d_i$ is the min distance to the support surface and $\mathrm{Hull}_i$ is
the contact-point convex hull.

\paragraph{Accessibility (ACC).}
A VLM identifies each object's \emph{functional sides} (e.g.\ the front of a
chair, the opening of a cabinet). For each functional side we cast rays from a
ring around the object toward it and measure the fraction that reaches the side
unblocked, giving a per-side score $s_{i,s}\in[0,1]$. An object's accessibility
is its best side, $a_i = \max_s s_{i,s}$; objects with no functional side are
excluded. The metric is the mean over the $N'$ accessible objects,
\begin{equation}
\mathrm{ACC} \;=\; \frac{1}{N'}\sum_{i=1}^{N'} \max_{s} s_{i,s}.
\end{equation}

\subsection{Prompt fidelity (spec satisfaction)}
\label{sec:metrics_spec}

A shared set of requirement specifications is derived from each prompt and a VLM-based \emph{object matcher} first assigns
each scene object to the target categories (yielding per-category groups
$\mathcal{C}$). Each metric below is the fraction of its requirements that are
satisfied,
\begin{equation}
\mathrm{Metric} \;=\; \frac{1}{N_{\text{spec}}}\sum_{j=1}^{N_{\text{spec}}} \mathbf{1}[\,\text{requirement } j \text{ holds}\,].
\end{equation}

\paragraph{Object count (CNT).}
Each requirement is $(q, n, c)$: ``the scene should contain $n$ objects of
category $c$ under quantifier $q$'' ($q\in\{\mathrm{eq,lt,gt,le,ge}\}$). It is
satisfied when the matched count $\#\mathcal{C}[c]$ satisfies the comparison,
e.g.\ for $q=\mathrm{eq}$, $\#\mathcal{C}[c] = n$.

\paragraph{Object attribute (ATR).}
Each requirement is $(q, n, c, a)$: ``at least/at most $n$ objects of category
$c$ should exhibit attribute $a$'' (material, shape, color). A VLM inspects the
rendered images of each matched object in $\mathcal{C}[c]$ and the requirement
holds when the number of objects judged to have attribute $a$ satisfies the
quantifier.

\paragraph{Object--object relationship (OO-Rel).}
Each requirement is $(q, n, r, \text{anchor}, \text{refs})$: ``$n$ target
objects stand in relation $r$ to an anchor object.'' The natural-language
relation $r$ (e.g.\ \emph{around}, \emph{next to}, \emph{on top of}) is mapped
by a VLM to canonical spatial types (inside\_of, on\_top, side\_of, surround,
face\_to, \dots), then verified geometrically against object bounding boxes and
poses.

\paragraph{Object--architecture relationship (OA-Rel).}
Each requirement is $(q, n, r, o, e)$: ``$n$ objects $o$ stand in relation $r$
to an architectural element $e$'' ($e\in\{\text{wall, floor, ceiling, window,
door}\}$, e.g.\ a sideboard \emph{against} the wall, a chandelier \emph{hanging
from} the ceiling). Verified geometrically from the object and architecture
bounding boxes.

\begin{table*}[h]
\centering
\small

\begin{tabular}{lll}
\toprule
Family & Metrics & What it measures \\
\midrule
Physical plausibility & COL, NAV, OOB, OPC & realistic, walkable, in-room layout \\
VLM-judged physical   & SUP, ACC & objects properly supported \& reachable \\
Prompt fidelity       & CNT, ATR, OO-Rel, OA-Rel & scene matches the text prompt \\
\bottomrule
\end{tabular}
\caption{The ten evaluation metrics and their families.}
\label{tab:metric_families}
\end{table*}

\section{D. Interactive Object Demonstrations}
\label{interact_obj}

Complex interaction in our scenes is \emph{code-based}: every interaction is
an explicit, machine-readable edge in a typed control topology, and the same
topology that drives asset synthesis is executed verbatim at simulation time.

\paragraph{Interaction topology inference.}
From the placed object roster, a single LLM pass infers three coordinated
structures. \emph{Controllers} are objects whose role is to initiate
interaction (wall switches, remotes, control panels). \emph{Part contracts}
declare the interactive parts each object must physically possess (buttons,
knobs, displays, sensors); contracts are passed to the articulation backend,
so every contract part exists in the final asset as a named URDF link with a
proper joint --- interaction capability is built into the geometry, not
asserted afterwards. \emph{Typed edges} wire controller parts to affected
parts using an \texttt{obj\#part} addressing scheme:

\begin{lstlisting}[caption={A part-level edge. Cross-object
edges use the identical schema with a different object id in target.},
label={lst:edge}]
{
  "source":   "obj_007#power_button",
  "target":   "obj_007#display",
  "relation": "activates",
  "method":   "binary_switch",
  "effect":   "screen",
  "reason":   "The power button toggles the microwave display and standby state."
}
\end{lstlisting}

The \texttt{method} field is drawn from a five-word vocabulary covering
\emph{how} control is applied --- \texttt{binary\_switch},
\texttt{multi\_knob}, \texttt{binary\_knob},
\texttt{binary\_remote\_button}, \texttt{contact} --- while
\texttt{effect} names the affected channel: \texttt{light},
\texttt{screen}, or \texttt{state}. Granularity is valve-controlled by the
model: complex appliances expose part-level edges (button $\rightarrow$
display), while simple devices collapse to object-level edges (switch
$\rightarrow$ lamp). Note that the 23 scenes analyzed here are those with
\emph{easy} and \emph{medium} difficulty prompts; harder prompts that demand
denser or more unusual control topologies are held out and not included in
these statistics.

\begin{figure}[t]
  \centering
   \includegraphics[width=1\linewidth]{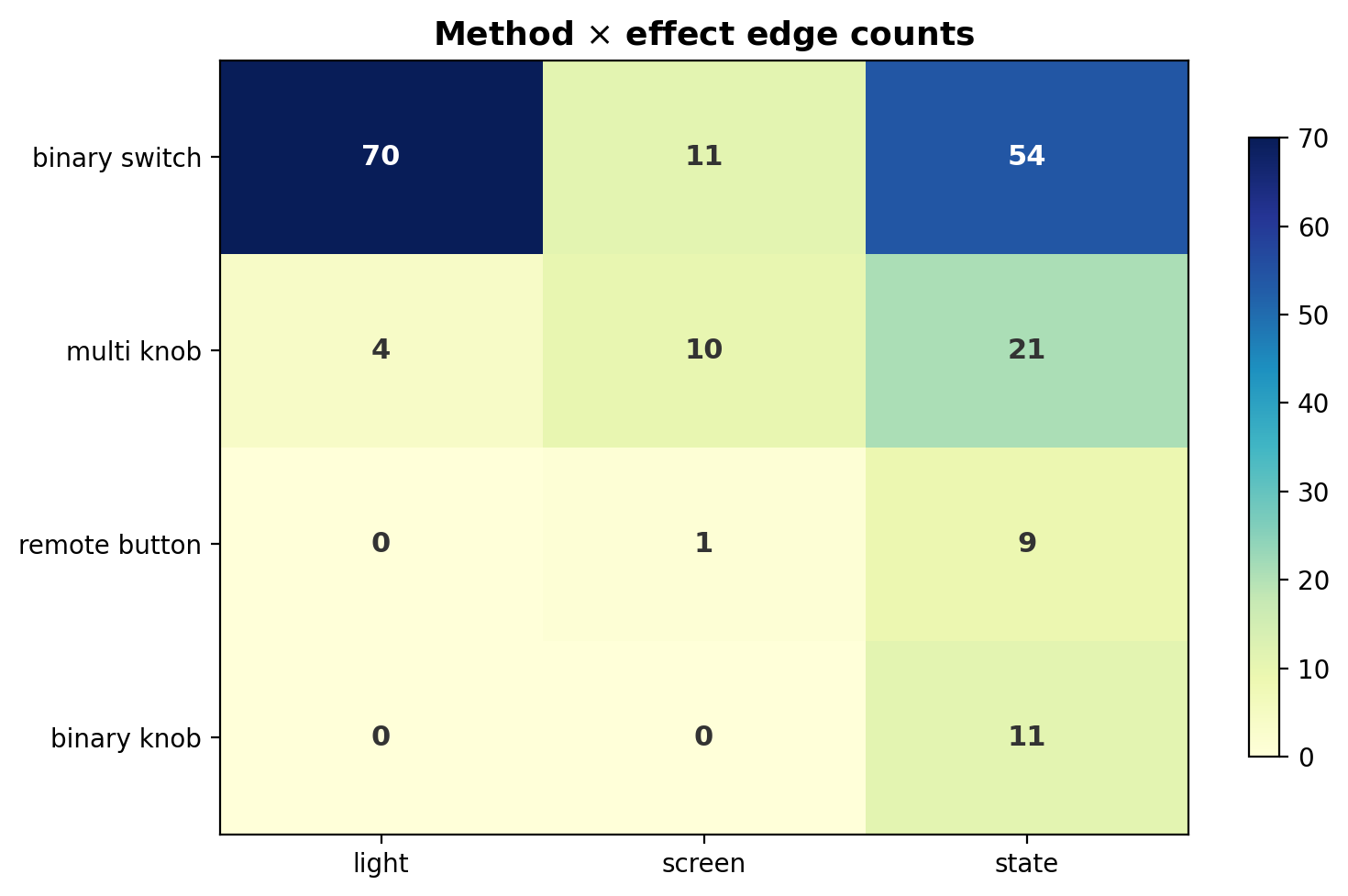}

   \caption{
   Method × effect distribution of the non-contact interaction edges across 23 scenes.
}
   \label{fig_inter_heat}
\end{figure}

Figure~\ref{fig_inter_heat} shows the resulting method $\times$ effect
matrix. The topology is typed but not degenerate: \texttt{binary\_switch}
dominates as expected for household control, including multi-state knobs driving screens and remote buttons driving state changes. The
\texttt{contact} method (e.g.\ proximity-triggered faucets)
completes the picture of hands-free interaction.

\begin{figure}[t]
  \centering
   \includegraphics[width=0.9\linewidth]{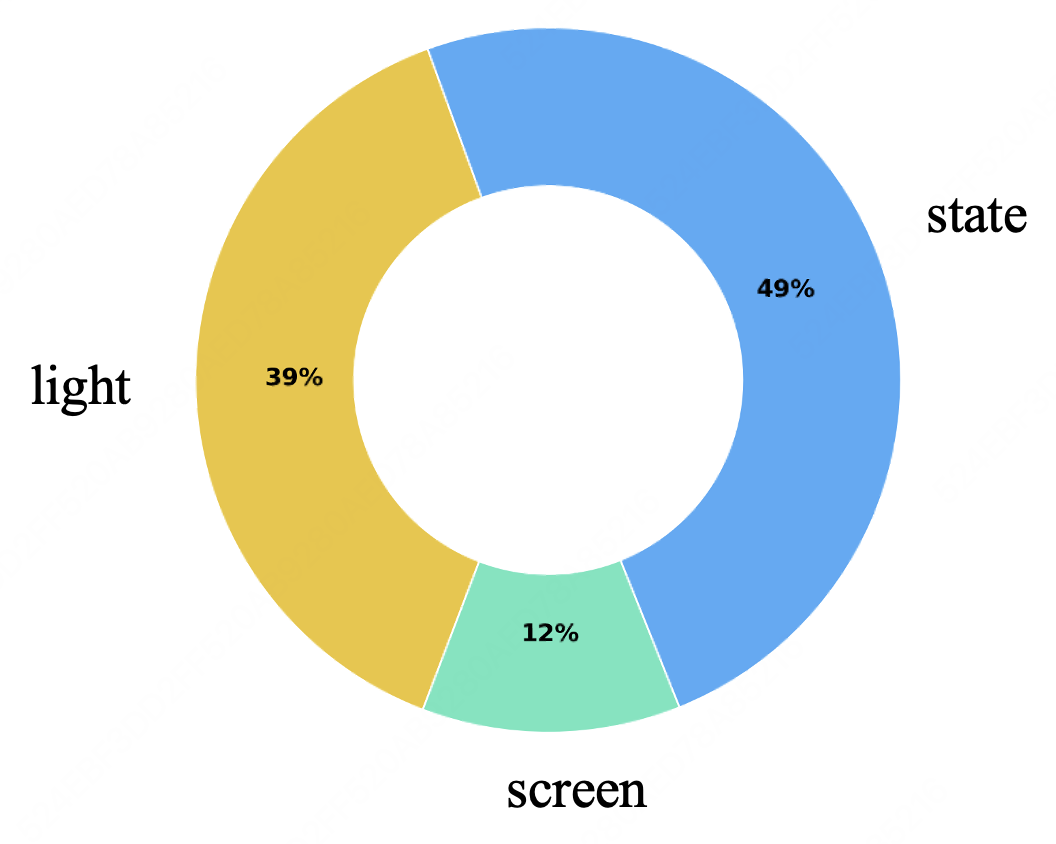}

   \caption{
   Effect-channel share over all inferred edges.
}
   \label{fig_inter_share}
\end{figure}

Figure~\ref{fig_inter_share} breaks the same edges down by effect channel.
Nearly half of all interactions target a discrete
\texttt{state} (a door position, a program, a power state) rather than a
luminous channel --- evidence that the topology captures functional,
mechanism-level interaction and not just switch $\rightarrow$ light wiring.

\begin{figure}[t]
  \centering
   \includegraphics[width=1\linewidth]{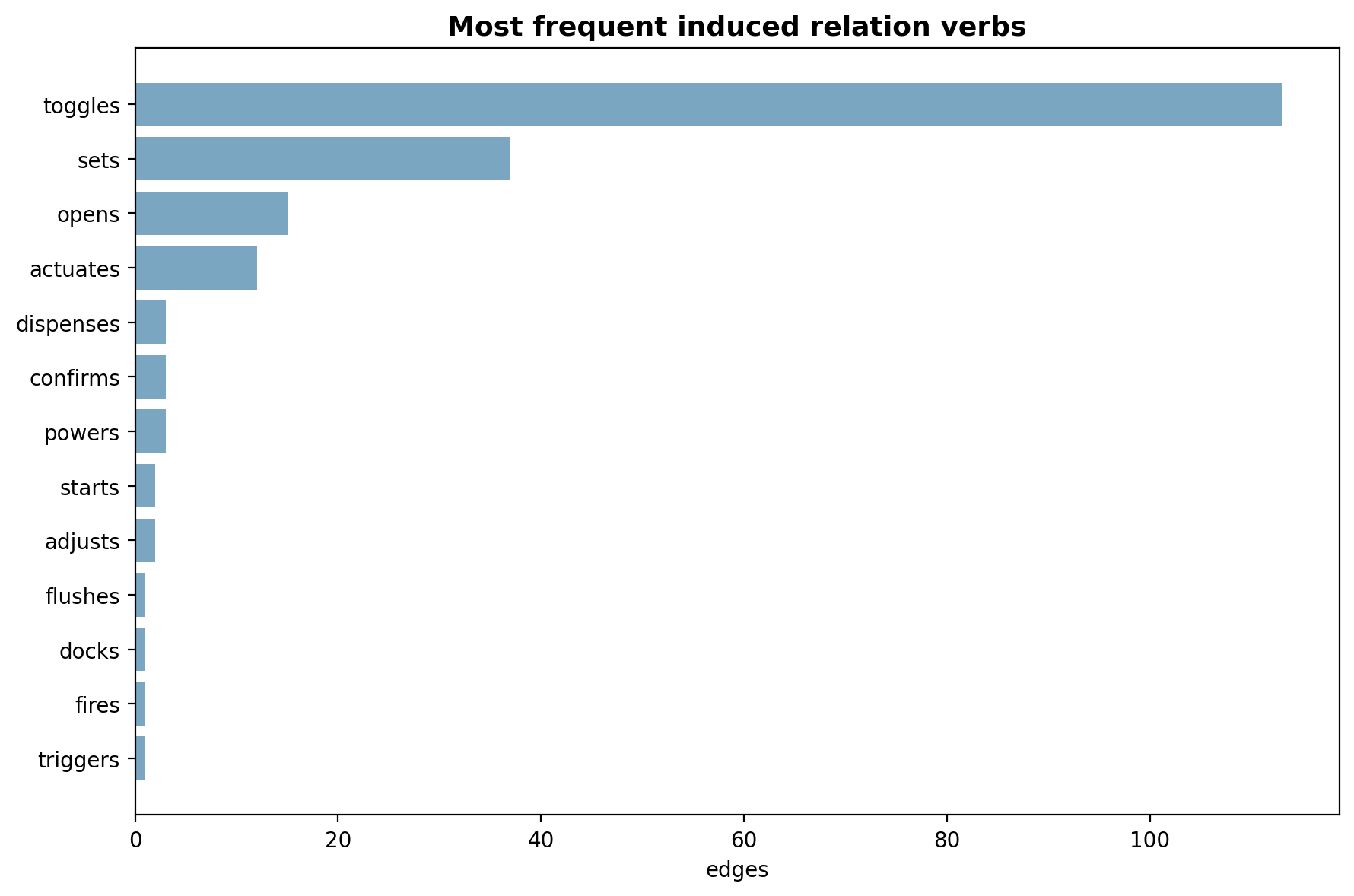}

   \caption{
   Most frequent induced relation verbs over the 194 edges.
}
   \label{fig_inter_verbs}
\end{figure}

Finally, Figure~\ref{fig_inter_verbs} shows that \texttt{relation} is an
\emph{open} vocabulary, not an enum: \texttt{toggles} and \texttt{sets}
cover the common cases, but the long tail contains scene-specific verbs that
the model induces from context (\texttt{opens} (doors, lids),
\texttt{dispenses} (soap, towels), \texttt{docks} (charging),
\texttt{flushes}, \texttt{fires}). These verbs are informative for downstream
consumers: the runtime resolves an edge by its method/effect pair, while the
relation verb documents \emph{intent} for search, debugging, and human
inspection.

\paragraph{Data-driven runtime.}
The same JSON is executed by a generic interpreter --- no object category
appears anywhere in runtime code. Each asset ships a small manifest mapping
contract parts to simulator prims and each part's effect to a material or
joint action; the manifest is generated from the edge's method/effect pair,
not hand-written per object:

\begin{lstlisting}[float=*, caption={Per-asset runtime manifest for the microwave edge
of Listing~\ref{lst:edge}.}, label={lst:manifest}]
{"parts": {
  "power_button": {"prim": "/World/microwave/power_button",
                   "joint": "body_to_power_button",
                   "press_threshold": 0.004},
  "display":      {"prim": "/World/microwave/display",
                   "effect": {"channel": "screen",
                              "on":  {"emissive_strength": 1.0},
                              "off": {"emissive_strength": 0.0,
                                      "albedo_dim": 0.15}}}}}
\end{lstlisting}

The interpreter loop is a few lines shared by every edge in every scene:

\begin{lstlisting}[caption={The entire category-agnostic interpreter.},
label={lst:runtime}]
for edge in topology.edges:
    src = resolve(edge.source)          # obj#part -> joint state
    tgt = resolve(edge.target)          # obj#part -> prim + effect bundle
    if joint_travel(src) >= src.press_threshold:
        apply_effect(tgt, ON)           # else apply_effect(tgt, OFF)
\end{lstlisting}

\begin{figure*}[t]
  \centering
   \includegraphics[width=1\linewidth]{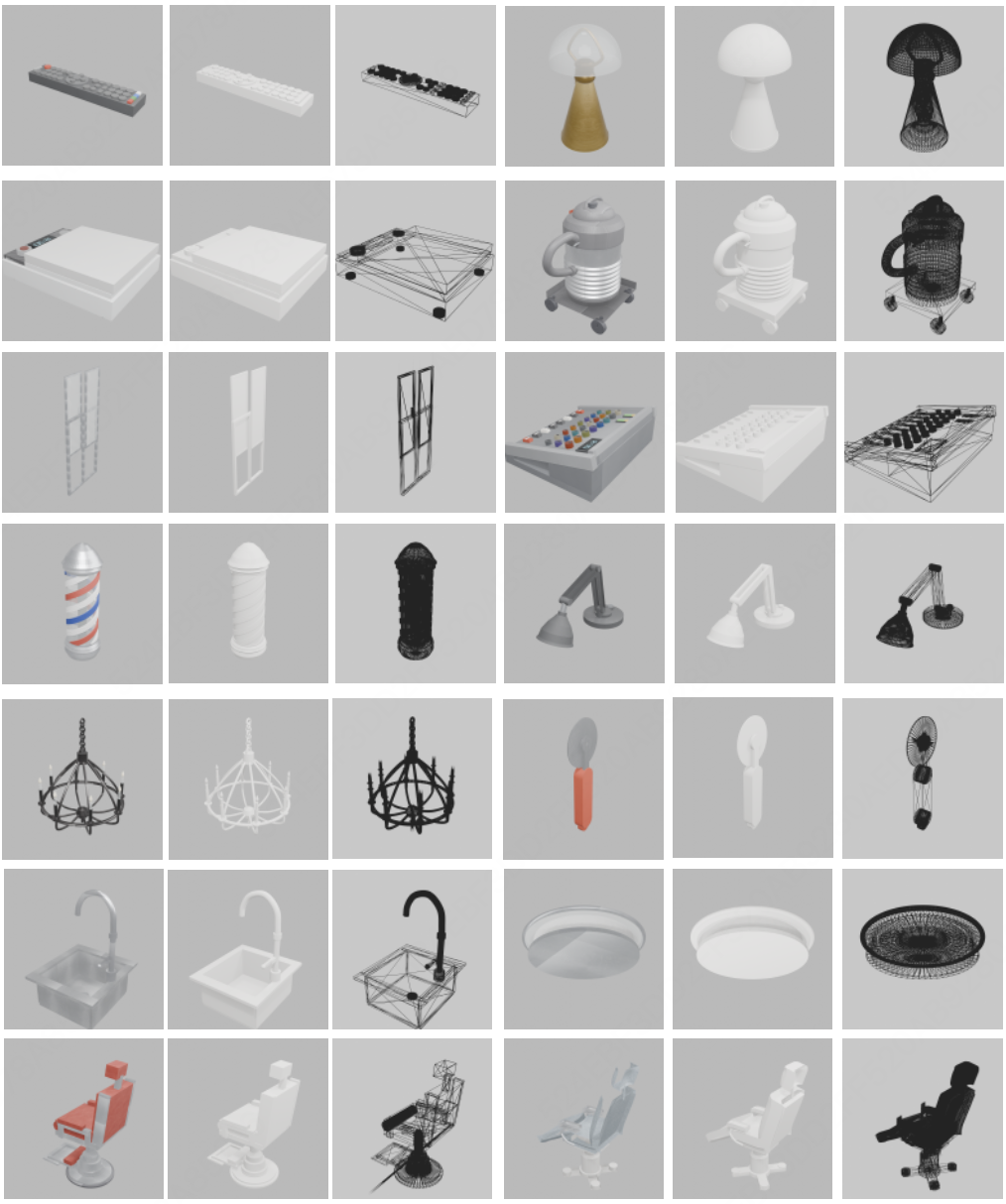}

   \caption{
   Demo of Objects with Complex Interactions
}
   \label{fig_drawer_demo}
\end{figure*}

Cross-object and part-level edges traverse the identical code path, because
addressing is object-scoped from the start: a wall switch toggling a ceiling
light and a button toggling its own display differ only in the manifest
entries their parts resolve to. Effects are material operations the
simulator natively supports (OmniPBR emissive toggling for
\texttt{screen}/\texttt{light} channels, with the off-state albedo dimmed so
unlit screens read correctly; joint articulation for \texttt{state}).
Figure~\ref{fig_drawer_demo} shows a gallery of the resulting interactive
assets --- each with its interactive parts realized as named, jointed links
--- in textured, untextured, and wireframe views. 
Finally, the architecture is extensible rather than fixed: adding a novel interaction means declaring a new edge, introducing new part contracts, method or effect vocabulary, and a manifest entry, allowing arbitrarily complex, user-defined control topologies to be layered onto the same runtime.

\section{E. Editability of Code-Represented 3D Assets}

Every object \method emits is a short \emph{program}, not an opaque mesh:
parameters control global shape (carcass width/height/depth, panel thickness),
and a handful of typed, parametric parts (\texttt{DrawerSlide},
\texttt{HingedDoor}, \texttt{CabinetCarcass}) are instantiated and placed by
plain code. This choice buys a property that mesh or
point-cloud-based scene generators structurally lack: \textbf{editability}.
We argue that editability shows up at four levels, all of which are exercised
in our pipeline and in the accompanying asset library.

\begin{figure*}[t]
  \centering
   \includegraphics[width=1\linewidth]{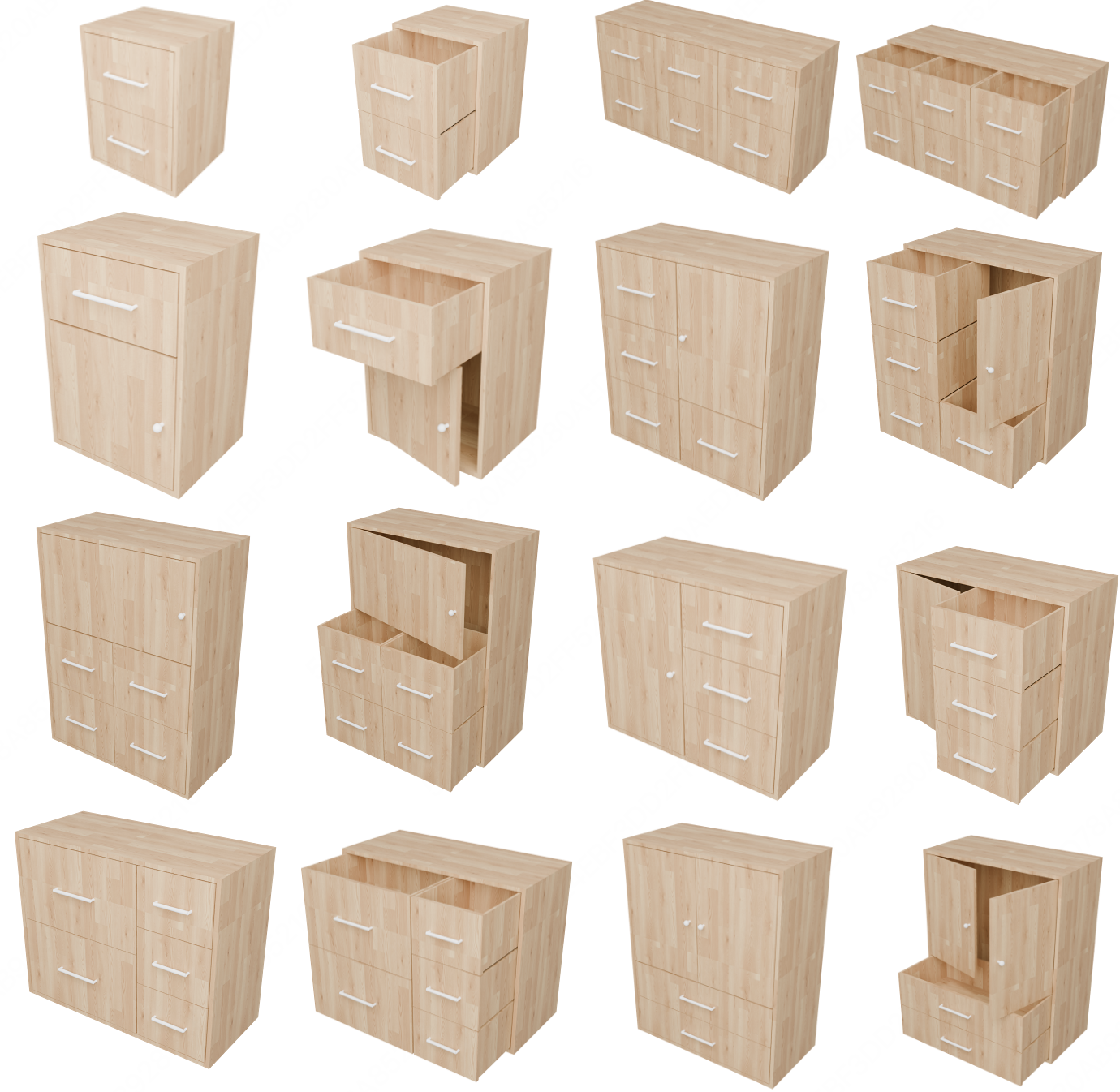}

   \caption{
   Parametric Editing of Code-Represented 3D Assets
}
   \label{fig_drawer_open}
\end{figure*}

\paragraph{(1) Parametric edits: one template, a family of assets.}
Because the asset is code, changing a few numbers yields a \emph{structurally
distinct} object while every geometric consistency rule is re-enforced by
construction. As a demonstration, a single drawer-cabinet template
(Listing~\ref{lst:drawer_template}) produced a library of storage
variants---vertical stacks, grids, and mixed door+drawer layouts; eight of
them are shown in Figure~\ref{fig_drawer_open} and enumerated in
Table~\ref{tab:drawer_variants}. The \texttt{rows}$\times$\texttt{cols} array
and the region splits are the \emph{only} degrees of freedom that change;
drawer fitment inside the carcass (gaps, depth, slide travel) is recomputed
each time, so no edit can produce a drawer that intersects its case. The
equivalent edit on a generated mesh would require re-modeling and re-rigging
each variant by hand.

\begin{lstlisting}[language=Python,float=t,
  caption={One parametric template spans the whole variant family. Only the
  layout lines differ between variants; part invariants are enforced by the
  typed parts.},
  label={lst:drawer_template}]
carcass = CabinetCarcass(width=0.9, height=1.0, depth=0.45)
case = carcass.attach_to(m, name="case", material=wood)
# grid: rows x cols drawers in region [x0,x1]x[z0,z1]
add_drawers(xi0, xi1, zi0, zi1, rows=3, cols=2, prefix="drawer")
# ... or a door: hinge edge x, height range [z0,z1]
add_door(xi0 + gap, zsplit, zi1, width=half, hinge="left", name="door_l")
\end{lstlisting}

\begin{table*}[t]
  \centering
  \small
  
  \begin{tabular}{ll}
    \toprule
     Layout parameters & Resulting asset \\
    \midrule
    \texttt{rows=2, cols=1} & nightstand \\
     \texttt{rows=2, cols=3} & wide dresser \\
     double doors top + \texttt{rows=2} bottom & cupboard with drawers \\
    \texttt{rows=1} top + door bottom & nightstand with cabinet \\
     door column + \texttt{rows=3} column & wardrobe with drawers \\
    \texttt{rows=3} column + door-over-drawer column & combo dresser \\
     asymmetric split (\texttt{2} wide + \texttt{3} narrow) & desk pedestal \\
    door top + \texttt{rows=2, cols=2} bottom & tall cupboard \\
    \bottomrule

  \end{tabular}
      \caption{The eight cabinet variants shown in Figure~\ref{fig_drawer_open},
  all generated from the single template of
  Listing~\ref{lst:drawer_template}. Each row is a parameter change.}
  \label{tab:drawer_variants}
\end{table*}

\paragraph{(2) Semantic edits: parts stay physically valid.}
Code-level edits act on \emph{named, typed} parts, so the asset's functional
semantics survive editing. A \texttt{DrawerSlide} always compiles to a
prismatic joint with the correct axis, parent, and travel limit (default
$0.9\times$ depth); a \texttt{HingedDoor} always rotates about its edge.
Swapping a drawer for a door, mirroring a hinge (\texttt{left}
$\rightarrow$ \texttt{right}), or re-styling a handle is a one-line change
that cannot silently break the joint graph---the invariants are locked in the
part implementation, not delegated to the generative model or the human
editor. In contrast, editing a mesh asset's articulation means re-assigning
joint frames by hand, a well-known source of broken assets.

\paragraph{(3) State edits: posing is free.}
Because articulation is explicit, any configuration state is obtained by
writing joint values, not by re-generating geometry. From each cabinet's
URDF we produced both the closed state and an \emph{open} state (drawers at
75\% of travel, doors at 85\% of their angular limit) by simply setting
$\mathbf{q}$ and re-evaluating the kinematics
(Figure~\ref{fig_drawer_open}). This is precisely what downstream consumers
need: a physics simulator can play the same joint trajectories, and a
renderer can depict the same object in any pose---the mesh baked at one
configuration cannot.

\paragraph{(4) Text-level edits: diffable, reviewable, agent-editable.}
Finally, a program is text. Variants are diffs, review is reading, and the
editing agent can be a human \emph{or an LLM}: the model that authors the
asset can also revise it (``make the left column a door instead''),
recompile, and re-render in a closed loop. Material and texture changes are
equally cheap---the cabinet library is re-skinned (oak albedo, white
handles) without touching geometry, since appearance is decoupled from the
code that defines shape and function.

\paragraph{Why this matters.}
Scene generators that output meshes entangle shape, articulation, and
appearance into an opaque artifact: any change costs a full re-generation
with no guarantee that structure is preserved. Code output makes the three
orthogonal---parameters reshape, joint values repose, materials reskin---
which is what allows our pipeline to \emph{iterate}: completion can add
objects, the critic can move them, and users can customize assets, all
without ever breaking the functional contract each object carries.

\label{edit}

\section{F. Computational Cost Statistics}
\label{cost}

We report the computational cost statistics for the same batch of evaluation runs used in our experiments. Object code is generated with Claude Opus 4.8, whereas the overall scene agent uses GPT-5.4. Table~\ref{tab:cost-tokens} summarizes token cost, and Table~\ref{tab:cost-time} summarizes wall-clock runtime. The reported cost covers both object code generation and the orchestration of the overall scene agent.

\begin{table}[h]
\centering\small
\caption{Token consumption statistics for the evaluation runs. }
\label{tab:cost-tokens}
\begin{tabular}{@{}lr@{}}
\toprule
Cost item & Tokens (M)\\
\midrule
Total token count & 1,204.5\\
Average token count & 43.0\\
Maximum token count & 72.0\\
Minimum token count & 27.2\\
\bottomrule
\end{tabular}
\end{table}

\begin{table}[h]
\centering\small
\caption{Wall-clock time cost statistics for the evaluation runs, reported both in hours:minutes:seconds and seconds.}
\label{tab:cost-time}
\begin{tabular}{@{}lrr@{}}
\toprule
Statistic & Wall-clock time & Seconds\\
\midrule
Average time cost & 9:10:38 & 33,038.0 s\\
Maximum time cost & 12:21:11 & 44,471.0 s\\
Minimum time cost & 6:34:49 & 23,689.0 s\\
\bottomrule
\end{tabular}
\end{table}

\section{G. Limitation}
\label{lim}

\begin{figure*}[t]
  \centering
   \includegraphics[width=1\linewidth]{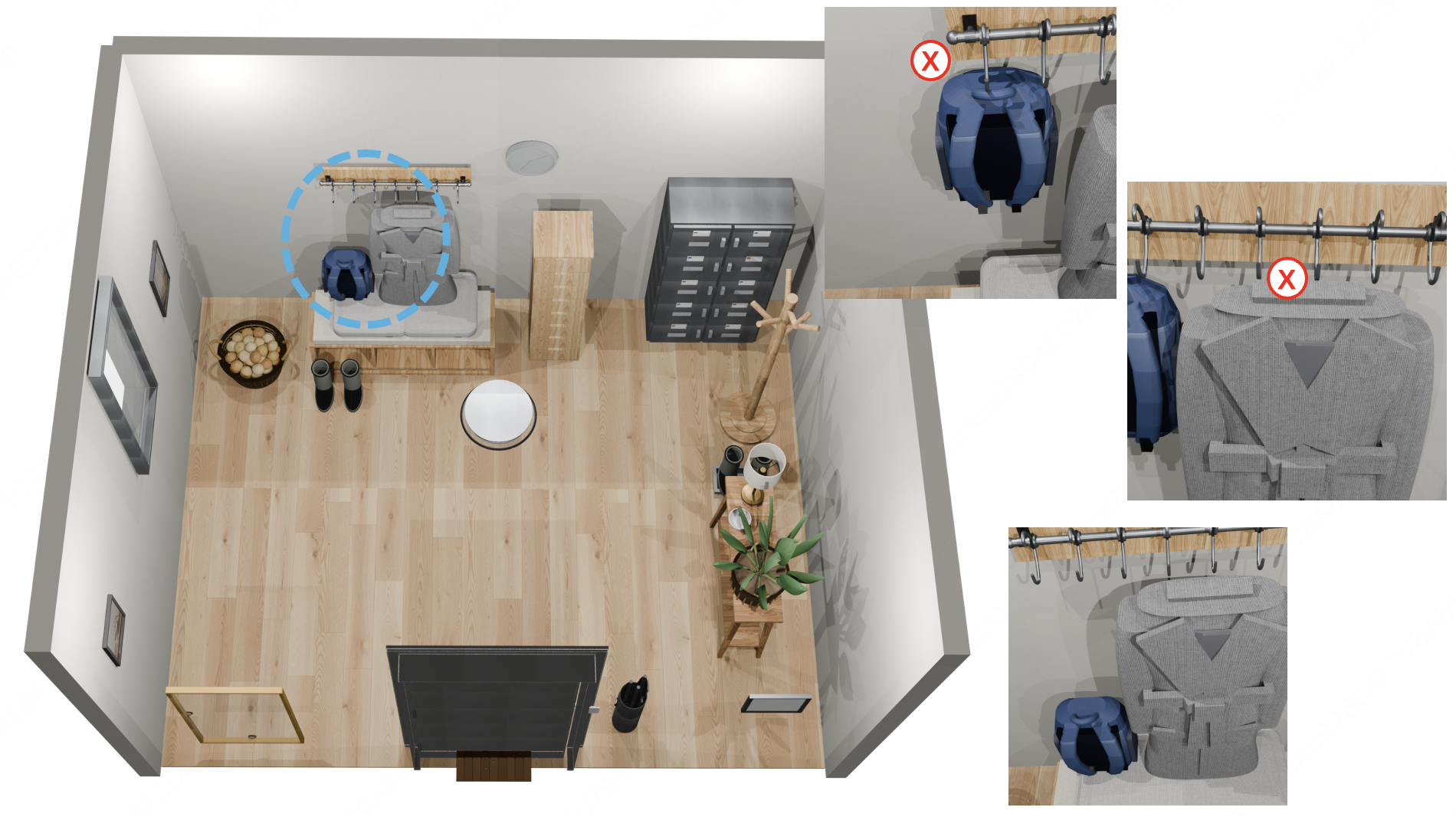}

   \caption{
   Failure Cases in Generated Scenes.
}
\label{fig_lim}
\end{figure*}

As shown in Figure~\ref{fig_lim}, objects that genuinely hang on hooks involve complex mechanical simulation, and the generated assets often lack suitable hanging structures. Moreover, our assets do not yet support soft cloth physics. As a result, such truly hanging objects cannot be realized in our scenes and degrade into ordinary placements.

\end{document}